\documentclass[10pt,twocolumn,letterpaper]{article}

\usepackage[pagenumbers]{cvpr_report} 

\definecolor{cvprblue}{rgb}{0.21,0.49,0.74}
\usepackage[pagebackref,breaklinks,colorlinks,allcolors=cvprblue]{hyperref}

\usepackage[table]{xcolor}
\usepackage{bm}

\def\paperID{13158} 
\def\confName{CVPR}
\def\confYear{2026}

\title{Reasoning Palette: Modulating Reasoning via Latent Contextualization\\ for Controllable Exploration for (V)LMs}

\author{
    Rujiao Long\textsuperscript{1}\thanks{Equal Contribution.} \quad 
    Yang Li\textsuperscript{1,2}\footnotemark[1] \quad 
    Xingyao Zhang\textsuperscript{1} \quad 
    Weixun Wang\textsuperscript{1} \quad 
    Tianqianjin Lin\textsuperscript{1,3}\\
    Xi Zhao\textsuperscript{1} \quad 
    Yuchi Xu\textsuperscript{1} \quad 
    Wenbo Su\textsuperscript{1} \quad 
    Junchi Yan\textsuperscript{2}\thanks{Correspondence Author.} \quad 
    Bo Zheng\textsuperscript{1}\footnotemark[2] \\ 
    {\normalsize
    \textsuperscript{1}Alibaba Group \qquad
    \textsuperscript{2}Shanghai Jiao Tong University \qquad
    \textsuperscript{3}Zhejiang University
    }
}

\begin{document}
\maketitle

\begin{abstract}
Exploration capacity shapes both inference‑time performance and reinforcement learning (RL) training for large (vision-) language models, as stochastic sampling often yields redundant reasoning paths with little high-level diversity. This paper proposes Reasoning Palette, a novel latent-modulation framework that endows the model with a stochastic latent variable for strategic contextualization, guiding its internal planning prior to token generation. This latent context is inferred from the mean-pooled embedding of a question–answer pair via a variational autoencoder (VAE), where each sampled latent potentially encodes a distinct reasoning context. During inference, a sampled latent is decoded into learnable token prefixes and prepended to the input prompt, modulating the model's internal reasoning trajectory. \textbf{In this way, the model performs internal sampling over reasoning strategies prior to output generation, which shapes the style and structure of the entire response sequence.}
A brief supervised fine-tuning (SFT) warm-up phase allows the model to adapt to this latent conditioning. Within RL optimization, Reasoning Palette facilitates structured exploration by enabling on-demand injection for diverse reasoning modes, significantly enhancing exploration efficiency and sustained learning capability. Experiments across multiple reasoning benchmarks demonstrate that our method enables interpretable and controllable control over the (vision-) language model's strategic behavior, thereby achieving consistent performance gains over standard RL methods.
\end{abstract}

\section{Introduction}
\label{sec:intro}

\begin{figure}
  \centering
    \includegraphics[width=1\linewidth]{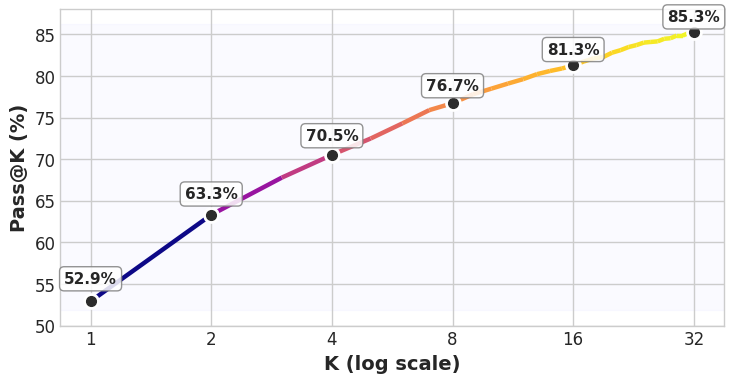}
    \vspace{-20pt}
    \caption{Motivation: Injecting a Gaussian noise token embedding before the prompt embeddings of Qwen-4B-Base enables substantial gains in pass@k accuracy by merely sampling in the Gaussian, despite using greedy decoding for each candidate.}
    \label{fig:motivation}
    \vspace{-10pt}
\end{figure}

\begin{figure*}
  \centering
  \begin{subfigure}{0.53\linewidth}
    \centering
    \includegraphics[width=0.9\linewidth]{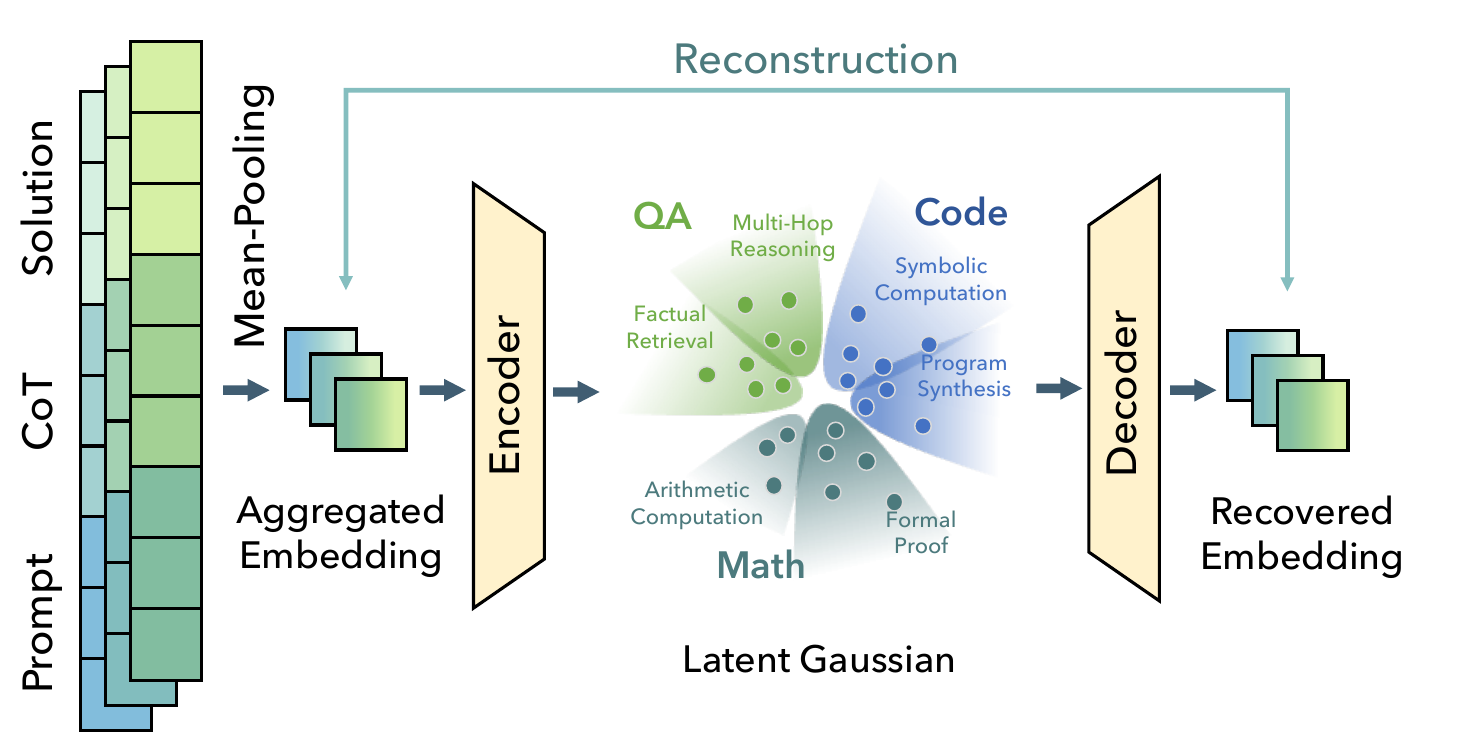}
    \caption{\textit{Latent space learning via VAE.} The encoder maps the mean-pooled embedding of a question–answer pair into a Gaussian latent space, where distinct regions correspond to different reasoning strategies. The decoder reconstructs the embedding, enabling structured exploration through latent sampling.}
    \label{fig:short-a}
  \end{subfigure}
  \hfill
  \begin{subfigure}{0.45\linewidth}
    \centering
    \includegraphics[width=0.9\linewidth]{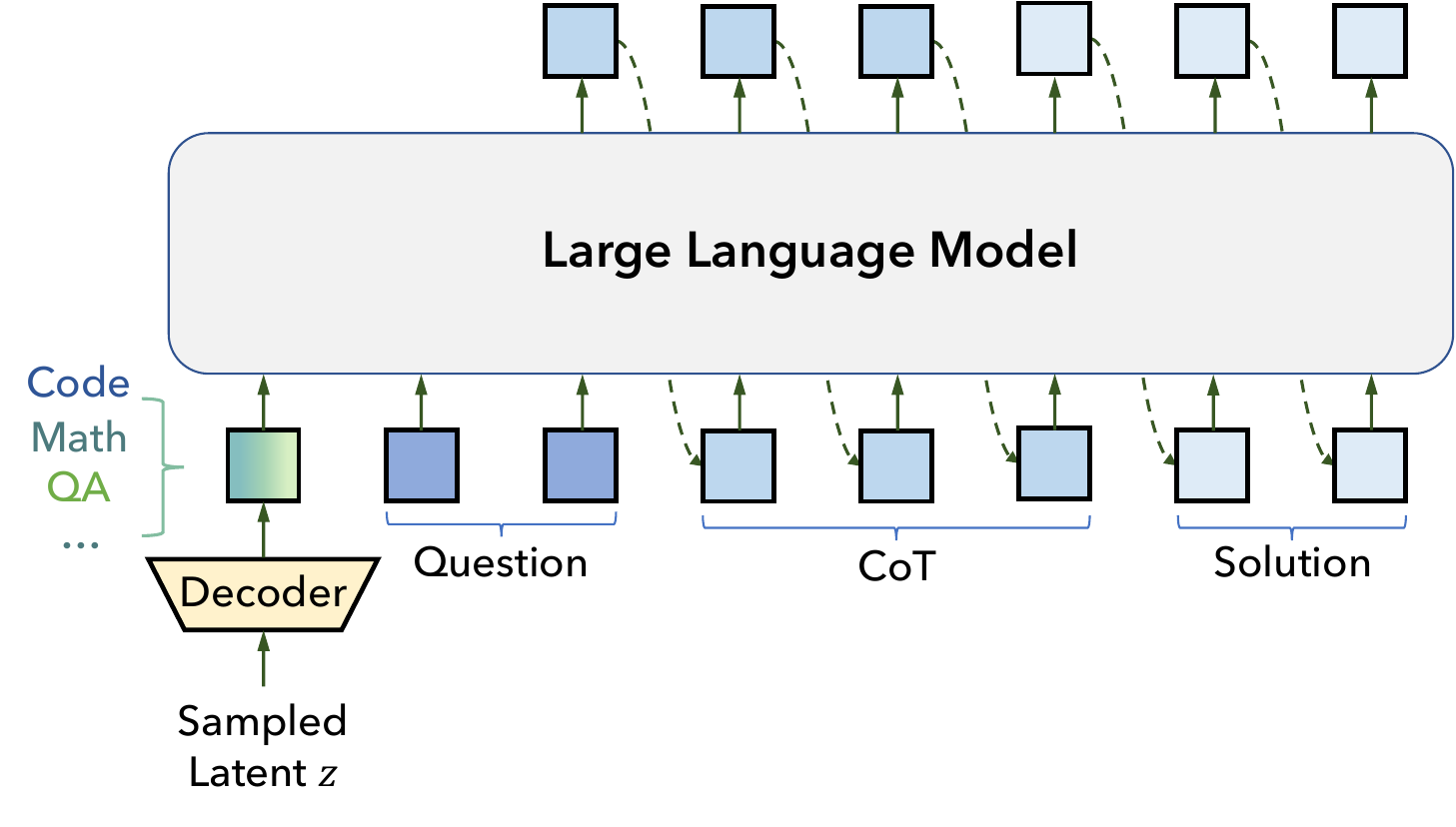}
    \caption{\textit{Latent-guided inference.} During generation, a sampled latent variable $z$ is decoded into prefixes and prepended to the input prompt. The prefixes modulate the LLM's internal state, steering its reasoning trajectory and shaping the style and structure of the generated response.}
    \label{fig:short-b}
  \end{subfigure}
  \vspace{-5pt}
  \caption{Overview of the Reasoning Palette framework: a latent-modulation system that enables strategic, diverse reasoning in LLMs/VLMs by sampling and decoding contextual latent variables to guide internal planning before token generation.}
  \label{fig:overview}
  \vspace{-10pt}
\end{figure*}

Large language models (LLMs) and vision-language models (VLMs) have rapidly advanced the frontier of artificial general intelligence, enabling multi-step problem solving, tool use, and grounded decision making across textual and multimodal domains~\cite{ghosh2024exploring,zhang2024vision,zhao2023survey,huang2023towards}. Recently, reinforcement learning with verifiable rewards (RLVR)~\cite{lambert2024tulu} has been widely verified as an effective post-training paradigm that elicits step-by-step thought for powerful reasoning, where a model's outputs are optimized with RL objectives guided by automated correctness checks. This setup encourages models to produce intermediate reasoning tokens before issuing a final answer, giving rise to Large Reasoning Models (LRMs) that excel on challenging problems in mathematics~\citep{hendrycks2021measuring,cobbe2021training,shao2024deepseekmath,yang2024qwen2}, coding~\citep{chen2021evaluating,jimenez2023swe,hui2024qwen2}, and agentic decision making~\citep{liu2023agentbench,yao2024tau}.

Exploration capacity shapes both inference‑time performance and reinforcement learning (RL) training for large (vision-) language models. This long-standing bottleneck in RL training becomes even more critical when it comes to language models: standard sampling schemes (e.g., stochastic decoding with temperature or nucleus sampling) often revisit closely related trajectories, resulting in limited diversity at the levels of strategy and plan structure. This mismatch, between the high-level variability required to discover effective reasoning strategies and the low-level variability of token-level sampling, limits the efficiency and robustness of RL training for LLMs. Existing remedies primarily encourage local diversity, e.g., entropy regularization~\cite{haarnoja2018soft,wang2025reinforcement}, but they rarely provide mechanisms to shape the model's internal planning. As a result, the model often explores similar reasoning paths with superficial differences, rather than probing distinct strategic modes.

Motivated by the phenomenon shown in Fig.~\ref{fig:motivation}, where prepending a single randomly sampled Gaussian noise embedding to the input of Qwen-4B-Base dramatically boosts Pass@$k$ performance even under greedy decoding, we propose \textbf{Reasoning Palette}, a latent modulation framework that endows the model with a compact stochastic variable acting as a latent context for internal planning, as shown in Fig.~\ref{fig:overview}. Concretely, we construct a latent space via a variational autoencoder (VAE), where each latent corresponds to a mean-pooled embedding of question-answer pairs. Each sampled latent is decoded into a short sequence of learnable token prefixes and prepended to the prompt during inference, thereby steering the model's internal trajectory before the first output token is produced. This design transforms exploration from token-level randomness into structured, pre-generative sampling over reasoning strategies. The latent context can be viewed as a palette of reasoning modes that can be explored through sampling, where different latent contexts inspire distinct organizational and explanatory styles of reasoning (e.g., mathematical reasoning, code reasoning, or question-answering logic).

A brief supervised fine-tuning (SFT) warm-up aligns the base model to respond to these latent-conditioned prefixes without sacrificing its general capabilities. Within RL, we treat the latent as an auxiliary control variable sampled per episode, enabling on-demand injection of diverse reasoning contexts. This facilitates efficient and sustained exploration: rather than repeatedly sampling near-duplicate chains of thought, the policy explores distinct strategic families, improving the chance of discovering high-reward behaviors and accelerating learning. Crucially, because the prefixes are generated by an external variational autoencoder (VAE), the reasoning mode is explicitly controllable. By identifying regions of the latent space associated with specific reasoning patterns (e.g., via encoding reasoning trajectories of targeted patterns), one can deliberately sample corresponding latents to modulate the model's behavior on new prompts. This enables both targeted intervention and systematic analysis of how different \textit{reasoning palettes} influence task success.

We summarize our contributions as follows: (1) We introduce a stochastic latent variable learned by VAE that modulates internal planning in LLMs and VLMs via token prefixes, enabling sampling over reasoning strategies before generation begins. (2) We propose a two-stage procedure to promote the reasoning capability, SFT warm-up to sensitize the model to latent conditioning, followed by RL with latent sampling to enhance exploration in optimization, which yields structured exploration and sustained learning. (3) Experiments on multiple reasoning benchmarks show that Reasoning Palette shows clear gains in diversity and controllability during inference, and consistently improves performance over standard baselines in RL, significantly enhancing exploration efficiency and controllability.

\section{Related Work}


\textbf{Controllable Reasoning and Prompting.} Chain-of-thought (CoT) prompting elicits step-by-step rationales that improve reasoning accuracy~\cite{wei2022chain}, and self-consistency enhances reliability through majority voting over diverse rationales~\cite{wang2022self}. Complementary paradigms structure the decomposition or search process~\cite{zhou2022least,yao2022react,yao2023tree}. Recently, soft chain-of-thought~\cite{xu2025softcot} replaces discrete reasoning traces with learnable continuous representations to boost efficiency. Beyond hand-crafted prompts, a large literature studies controllable generation via discrete and continuous prompt parameterizations, e.g., soft prompt tuning \citep{lester2021power} and prefix-tuning \citep{li2021prefix}, which learn continuous prefix embeddings that steer model behavior with minimal updates, or discrete prompt search methods~\citep{shin2020autoprompt,fernando2023promptbreeder}. Orthogonal control uses external guidance at inference, such as plug-and-play gradients and classifiers \citep{dathathri2019plug, krause2020gedi}, or activation steering to modulate internal representations~\cite{turner2023steering}.

\textbf{Reinforcement Learning for LLMs.} Reinforcement learning (RL) is central to LLM post-training, popularized by RLHF for instruction following and preference alignment~\citep{christiano2017deep, ziegler2019fine, ouyang2022training, schulman2022chatgpt, bai2022training}. Two strands dominate: online policy-gradient methods with on-policy rollouts~\citep{schulman2017proximal,shao2024deepseekmath,williams1992simple} and offline preference optimization without on-policy sampling~\citep{rafailov2023direct,meng2024simpo,ethayarajh2024kto}. 
Reinforcement Learning from AI Feedback (RLAiF) extends this by using LLMs to generate feedback for alignment~\citep{bai2022constitutionalaiharmlessnessai, sun2023principle, lee2023rlaif, guo2024directlanguagemodelalignment}.
Outcome-based RL with verifiable rewards (RLVR)~\citep{shao2024deepseekmath,lambert2025tulu3pushingfrontiers} has driven recent advances through deterministic rule-based rewards~\citep{xin2024deepseekproverv15harnessingproofassistant, wang2024mathshepherdverifyreinforcellms}, with large-scale systems demonstrating that correctness-guided RL elicits extended reasoning~\citep{lambert2024tulu,openai2024learning,guo2025deepseek,team2025kimi,yang2025qwen3}. A growing ecosystem studies data aggregation, policy updates, scalable training infrastructure, and critical reasoning nodes that enhance optimization~\citep{yu2025dapo,yue2025vapo,zheng2025groupsequencepolicyoptimization, liu2025part, wang2025beyond, li2025attention}. 


\textbf{Latent Representation Learning.} Learning compact and disentangled latent representations has been a longstanding theme in generative modeling. Canonical approaches include variational autoencoders (VAEs)~\cite{kingma2013auto} that optimize a tractable ELBO to learn continuous, sampleable latents, generative adversarial networks (GANs) that leverage adversarial formulations to align latent codes with data distributions, and diffusion models~\cite{ho2020denoising,yang2023diffusion} that construct latent-like representations by progressively denoising inputs across timesteps. Discrete codebooks (VQ-VAE)~\citep{van2017neural} provide robust, compositional latents with high reconstruction quality. For text, latent-variable language models capture global factors while mitigating posterior collapse via architectural and training advances \citep{bowman2016generating,he2019lagging}. More recently, iterative and multi-step generation strategies that preserve or refine intermediate latent embeddings have been explored to support multi-step reasoning and better align latent dynamics with downstream generation~\cite{hao2024training}.


\section{Methodology}

\subsection{Preliminaries and Notations}

Let $\mathbf{q}\sim Q$ denote a question, a decoder-only policy $\pi_\theta$ autoregressively generates an output token sequence $\mathbf{o}=(o_1, \cdots, o_T)$. When visual input $\mathbf{v}$ is present, we encode it with a visual encoder $f_v(\cdot)$ to obtain visual tokens
\begin{equation}
\mathbf{z} = f_v(\mathbf{v}) = (\mathbf{z}_1,\dots,\mathbf{z}_M),\quad \mathbf{z}_i\in\mathbb{R}^d,
\end{equation}
and the question $\mathbf{q}$ include the visual features. The autoregressive policy is formulated as:
\begin{equation}\label{eq:text_policy}
p_\theta(\mathbf{o}\mid\mathbf{q})
= \prod_{t=1}^T p_\theta\big(o_t \mid \mathbf{q},\mathbf{o}_{<t}\big).
\end{equation}

The RL objective maximizes expected verifier reward:
\begin{equation}
    \max_{\theta} \mathcal{J}(\theta) := \mathbb{E}_{\mathbf{q}\sim Q,\mathbf{o}\sim\pi_\theta(\cdot|\mathbf{q})}[r(\mathbf{o};\mathbf{q})].
\end{equation}
Proximal Policy Optimization (PPO)~\citep{schulman2017proximal} is a standard approach for optimization, which updates $\pi_\theta$ using data from a frozen old policy $\pi_{\mathrm{old}}$ and a clipped surrogate objective:
\begin{equation}\label{eq:ppo_loss}
\small
\begin{aligned}
\mathcal{J}(\theta) =&\mathbb{E}_{\mathbf{q}\sim Q,\mathbf{o}\sim\pi_\theta(\cdot|\mathbf{q})}
  \frac{1}{|\mathbf{o}|} \sum_{t=1}^{|\mathbf{o}|}
  \min\Bigg(
    \frac{\pi_\theta(o_t|\mathbf{q}, \mathbf{o}_{<t})}{\pi_{\theta_{\mathrm{old}}}(o_t|\mathbf{q}, \mathbf{o}_{<t})} A_t,\, \\
    &\mathrm{clip}\left(
      \frac{\pi_\theta(o_t|\mathbf{q}, \mathbf{o}_{<t})}{\pi_{\theta_{\mathrm{old}}}(o_t|\mathbf{q}, \mathbf{o}_{<t})},\, 1{-}\epsilon,\, 1{+}\epsilon
    \right) A_t
  \Bigg)
\end{aligned}
\end{equation}
where  $A_t$ is the advantage at step $t$, typically estimated via Generalized Advantage Estimation (GAE)~\citep{schulman2015high}, and $\epsilon$ is a clipping hyperparameter for stabilizing updates. Group Relative Policy Optimization (GRPO) ~\citep{shao2024deepseekmath} replaces the critic with group-based advantage estimation. For a prompt $\mathbf{q}$ with $G$ responses and rewards $\{r_i\}_{i=1}^G$, it computes $\hat{A}_{i,t} = \frac{r_i - \mathrm{mean}(\{r_i\}_{i=1}^G)}{\mathrm{std}(\{r_i\}_{i=1}^G)}$, enhancing gradient signal reliability under sparse rewards. GRPO further augments Eq.~\eqref{eq:ppo_loss} with a KL penalty to stabilize training.

\subsection{Latent Space Learning for Reasoning Patterns}
\label{subsec:vae}

We learn a compact, continuous latent space $\mathcal{Z} \subset \mathbb{R}^k$ that encodes diverse reasoning strategies. To construct this space, we train a Variational Autoencoder (VAE)~\citep{kingma2013auto} on a dataset $\mathcal{D} = \{(\mathbf{q}^{(i)}, \mathbf{o}^{(i)})\}_{i=1}^N$ of question–answer pairs with high-quality reasoning traces. To ensure diversity in reasoning patterns across different modes, we curate the training data from a broad spectrum of sources spanning mathematics, question answering, and code generation. Each of these domains exhibits distinct reasoning patterns, while also containing multiple sub-modes. For example, question answering encompasses strategies such as multi-hop reasoning, factual retrieval, and more. This intentional design ensures that the source data and consequently the learned latent space reflects a rich and varied repertoire of reasoning behaviors.

For each question–answer pair $(\mathbf{q}, \mathbf{o})$, we aim to learn a compact latent representation $\mathbf{z}$ that captures high-level reasoning strategies and can be flexibly decoded into a short prefix to guide generation. Critically, we desire two practical properties for downstream use: 
\begin{itemize}
    \item The number of prefix tokens $L$ is \textit{adjustable at inference time}, enabling users to adjust the control intensity through the length: longer prefixes provide stronger behavioral guidance and greater contextual diversity, while shorter prefixes offer lighter intervention.
    \item The prefix can be easily absorbed by the pretrained model without extensive retraining. Ideally, it resides in a space that is \textit{naturally aligned} with the model's native token embedding distribution, ensuring that the injected prefix harmonizes with the model's internal dynamics.
\end{itemize}

To satisfy these desiderata, we construct the latent space directly over the model's frozen token embedding layer $\mathcal{E}(\cdot)$ (e.g., from Qwen3-4B), which maps discrete tokens to vectors in $\mathbb{R}^d$. Specifically, we first form the concatenated sequence $[\mathbf{q}; \mathbf{o}] = (x_1, \dots, x_N)$, then compute its contextual summary via mean-pooling of the raw token embeddings:
\begin{equation}
\mathbf{h} = \frac{1}{N} \sum_{i=1}^{N} \mathcal{E}\big( [\mathbf{q}; \mathbf{o}]_i \big) \in \mathbb{R}^d.
\end{equation}
Since $\mathbf{h}$ lives in the same space as individual token embeddings, any reconstruction $\hat{\mathbf{h}}$ produced by the VAE decoder can be readily interpreted as a \textit{pseudo-token} or used to initialize a small set of prefix embeddings that are distributionally consistent with the model's input expectations. Moreover, mean-pooling provides a fixed-dimensional, order-insensitive summary that abstracts away surface-level token variations, making it well-suited for representing reasoning \textit{style} rather than verbatim output.

We then model the distribution over reasoning strategies by learning a probabilistic latent space using a Variational Autoencoder (VAE)~\citep{kingma2013auto}. The encoder $E_\phi$, implemented as a multi-layer perceptron (MLP), maps the embedding $\mathbf{h}$ to the parameters of a diagonal Gaussian distribution:
\begin{equation}
    \bm\mu, \bm\sigma = E_\phi(\mathbf{h}), \quad \mathbf{z} \sim \mathcal{N}(\bm\mu, \text{diag}(\bm\sigma^2)),
\end{equation}
where $\bm\mu \in \mathbb{R}^k$ and $\bm\sigma \in \mathbb{R}^k_{>0}$ define the posterior $q_\phi(z \mid \mathbf{h})$. The decoder $D_\psi$, also an MLP, reconstructs the original embedding from a sampled latent code:
\begin{equation}
    \hat{\mathbf{h}} = D_\psi(\mathbf{z}).
\end{equation}
The VAE is trained end-to-end by minimizing the evidence lower bound (ELBO), which balances reconstruction accuracy against latent space regularization:
\begin{equation}
    \mathcal{L}^{\text{VAE}} = \mathbb{E}_{\mathbf{z} \sim q_\phi(\mathbf{z} \mid \mathbf{h})} \big[ \|\mathbf{h} - \hat{\mathbf{h}}\|^2 \big] + \beta \cdot \text{KL}\big(q_\phi(\mathbf{z} \mid \mathbf{h}) \,\|\, p(\mathbf{z})\big),
\end{equation}
where $p(\mathbf{z}) = \mathcal{N}(0, I)$ is a standard isotropic Gaussian prior, and $\beta > 0$ is a hyperparameter that controls the trade-off between faithful reconstruction and smoothness of the latent manifold. A well-tuned $\beta$ encourages disentangled and semantically meaningful regions in $\mathcal{Z}$, such that nearby latents correspond to similar reasoning patterns, while distant latents induce qualitatively different strategies. 


\subsection{Latent-Guided Inference and SFT Adaptation}
\label{subsec:sft}

\textbf{Latent-Guided Inference.} Given a trained VAE $(E_\phi, D_\psi)$, we leverage its latent space to modulate the reasoning behavior of the base model during both inference and RL training. At inference time, we sample a latent vector $\mathbf{z} \sim \mathcal{N}(0, I)$ from the prior and decode it into a sequence of $L$ prefix embeddings:
\begin{equation}
    \mathbf{p}_\mathbf{z} = \big( D_\psi(\mathbf{z}^{(1)}), D_\psi(\mathbf{z}^{(2)}), \dots, D_\psi(\mathbf{z}^{(L)}) \big) \in \mathbb{R}^{L \times d},
\end{equation}
where each $\mathbf{z}^{(\ell)}$ is an independent sample from the same prior (or, optionally, a single $\mathbf{z}$ tiled $L$ times for stronger coherence). For inference, these continuous prefix embeddings $\mathbf{p}_\mathbf{z}$ are prepended to the input token embeddings of the prompt $\mathbf{q}$ for autoregressive generation:
\begin{equation}
    \tilde{\mathbf{q}} = [\mathbf{p}_\mathbf{z}; \mathcal{E}(\mathbf{q})], \quad \text{then} \quad \mathbf{o} \sim \pi_\theta(\cdot \mid \tilde{\mathbf{q}}).
\end{equation}

Crucially, the prefix length $L$ is a tunable hyperparameter: a larger $L$ provides stronger and more structured guidance (e.g., for complex multi-step tasks), while a smaller $L$ offers lightweight modulation with minimal computational overhead. This flexibility allows users to balance control intensity against inference cost. Moreover, since the VAE is fixed after SFT, the latent space remains a stable, interpretable coordinate system. This enables post-hoc analysis (e.g., clustering high-reward latents) and targeted intervention (e.g., biasing sampling toward known effective regions). Thus, Reasoning Palette not only improves exploration efficiency but also provides a transparent mechanism for steering and diagnosing model behavior.

\noindent\textbf{SFT Adaptation.} To ensure the base (V)LM can effectively interpret latent-conditioned prefixes without compromising its general instruction-following capabilities, we perform a lightweight supervised fine-tuning (SFT) warm-up phase. Since the learned posterior distribution $q_\phi(\mathbf{z} \mid \mathbf{h})$ can deviate from the isotropic Gaussian prior $p(\mathbf{z}) = \mathcal{N}(0, I)$, to avoid degraded generalization capability when using the latent Gaussian distribution during inference, we construct the SFT dataset $\mathcal{D}_{\text{SFT}}$ not by encoding ground-truth examples into latents, but by \textit{sampling latents directly from the prior} $\mathbf{z} \sim \mathcal{N}(0, I)$, decoding them into prefix embeddings $\mathbf{p} = D_\psi(\mathbf{z})$, and pairing these synthetic prefixes with original prompt–response pairs $(\mathbf{q}, \mathbf{o})$ from the VAE training corpus. 

We strictly limit the SFT duration (typically 10 iterations) to prevent the model from downweighting the impact of the prefix and overfitting to any single response pattern. Excessive SFT on deterministic targets can suppress the model's responsiveness to latent variation, effectively \textit{washing out} the diversity induced by different $\mathbf{z}$ samples. By keeping SFT minimal, we strike a balance: the model learns to condition its output on arbitrary prefix embeddings drawn from the VAE decoder's range, while retaining sufficient stochasticity to respond flexibly to novel latent codes.

The model is then fine-tuned to minimize the standard language modeling loss:
\begin{equation}
    \mathcal{L}_{\text{SFT}} = -\mathbb{E}_{(\mathbf{p}, \mathbf{q}, \mathbf{o}) \sim \mathcal{D}_{\text{SFT}}} \left[ \log p_\theta(\mathbf{o} \mid [\mathbf{p}; \mathcal{E}(\mathbf{q})]) \right],
\end{equation}
where $[\mathbf{p}; \mathcal{E}(\mathbf{q})]$ denotes the concatenation of prefix embeddings and question token embeddings. Notably, during SFT we fix the prefix length to $L = 1$, using only a single learnable token to modulate behavior. This minimal intervention enables the model to adapt to a noisy prefix token while avoiding overwhelming the model with strong control signals early on. During inference and RL, we freely increase $L$ (e.g., $L = 4$ or $8$) to enable richer, compositional guidance. This SFT phase sensitizes the model to the latent signal while preserving its pre-trained knowledge, enabling seamless integration of reasoning-mode control during subsequent RL.


\subsection{Controllable Exploration in RL}
\label{subsec:rl}

During reinforcement learning, we treat the latent variable $\mathbf{z}$ as an auxiliary control signal that induces structured exploration over high-level reasoning strategies. For each training episode, we sample a latent code $\mathbf{z} \sim \mathcal{N}(0, I)$, decode it into a sequence of $L$ prefix embeddings $\mathbf{p} = (D_\psi(\mathbf{z}^{(1)}), \dots, D_\psi(\mathbf{z}^{(L)}))$, and condition the policy on this prefix. The resulting policy becomes:
\begin{equation}
    \pi_\theta(\mathbf{o} \mid \mathbf{q}, \mathbf{z}) = \prod_{t=1}^T p_\theta\big(o_t \mid [\mathbf{p}; \mathcal{E}(\mathbf{q})], \mathbf{o}_{<t} \big),
\end{equation}
and the RL objective is extended to:
\begin{equation}
    \max_{\theta} \mathbb{E}_{\mathbf{z} \sim p(\mathbf{z})} \mathbb{E}_{\mathbf{q} \sim Q, \mathbf{o} \sim \pi_\theta(\cdot \mid \mathbf{q}, \mathbf{z})} \big[ r(\mathbf{o}; \mathbf{q}) \big].
\end{equation}
In practice, we approximate this by sampling one $z$ per prompt during rollouts, which efficiently diversifies reasoning trajectories across episodes.


This design enables \textit{multi-granular exploration control} in RL, operating along two complementary dimensions:  
(1) \textbf{temporal scheduling}, where the strength of latent guidance is modulated over the course of training to follow a high-level exploration-to-exploitation trajectory; and  
(2) \textbf{intra-group diversity control}, where, within each training batch (or GRPO group), we regulate the \textit{proportion} of rollouts that receive latent-conditioned prefixes, thereby fine-tuning the balance between diverse exploration and stable policy updates at the micro level. Based on this dual-axis framework, we evaluate two principled scheduling strategies:
\begin{itemize}[leftmargin=*]
    \item \textbf{Two-Phase}: For the first 50\% of training steps, all rollouts use latent prefixes with $L = 8$ to maximize strategic diversity; for the remaining steps, latent guidance is completely disabled ($L = 0$) across all samples, allowing the policy to fully exploit the high-reward behaviors discovered during exploration.
    \item \textbf{Linear Decay}: Over the full training duration, we linearly reduce the \textit{fraction of latent-guided rollouts per group} from 100\% to 0\%, while keeping $L = 8$ for those that are guided. This implements a smooth transition: early groups contain mostly diverse trajectories, while later groups increasingly rely on high-confidence generations for advantage estimation and policy refinement.
\end{itemize}

Both strategies embody the classical exploration-exploitation trade-off, but operate at the level of \textit{reasoning architecture} rather than token-level stochasticity. As shown in Fig.~\ref{fig:curves}, latent-augmented variants exhibit slower initial accuracy gains due to the cost of exploring diverse, sometimes suboptimal, reasoning paths, but ultimately surpass the GRPO baseline in the latter half of training. This late-stage overtaking occurs because early exploration exposes the policy to higher-quality regions of behavior space, which are then consolidated during the exploitation phase.

We formalize this time-varying control via a scheduled policy that mixes latent-guided and standard rollouts. Let $\tau \in [0,1]$ denote normalized training progress, and let $\rho(\tau) \in [0,1]$ be the scheduled fraction of rollouts that receive latent conditioning at step $\tau$. The optimization objective becomes a weighted combination:
\begin{equation}\label{eq:sched_rl}
\footnotesize
\mathcal{J}_{\text{sched}}(\theta) = \mathbb{E}_{\tau} \mathbb{E}_{\mathbf{q} \sim Q} \left[
    \rho(\tau) \cdot \mathcal{L}_{\text{PPO}}(\theta; \mathbf{q}, \mathbf{z}) +
    (1 - \rho(\tau)) \cdot \mathcal{L}_{\text{PPO}}(\theta; \mathbf{q})
\right],
\end{equation}
where $\rho(\tau)$ defines the exploration schedule. For the \textbf{two-phase} strategy, $\rho(\tau) = 1$ if $\tau < 0.5$ and $0$ otherwise; for \textbf{linear decay}, $\rho(\tau) = 1 - \tau$. In both cases, latent-guided rollouts use a fixed prefix length (e.g., $L = 8$), while unguided rollouts use $L = 0$. In practice, $\tau$ is determined deterministically by the current training step, and during each GRPO group rollout, we sample a fraction $\rho(\tau)$ of responses with latent prefixes and the remainder without.

\section{Experiments}
\label{sec:exp}

We conduct a comprehensive empirical study to validate the efficacy of Reasoning Palette across both inference and reinforcement learning settings. Our experiments are structured around two core questions: (1) Does latent-guided inference enable diverse and controllable reasoning? (2) Does integrating latent-conditioned exploration into RL improve training efficiency, stability, and final performance on challenging reasoning tasks? We evaluate primarily on mathematical reasoning benchmarks due to their clear reward structure and sensitivity to reasoning strategy diversity.

\begin{figure*}
  \centering
  \begin{subfigure}{0.23\linewidth}
    \centering
    \includegraphics[width=1\linewidth]{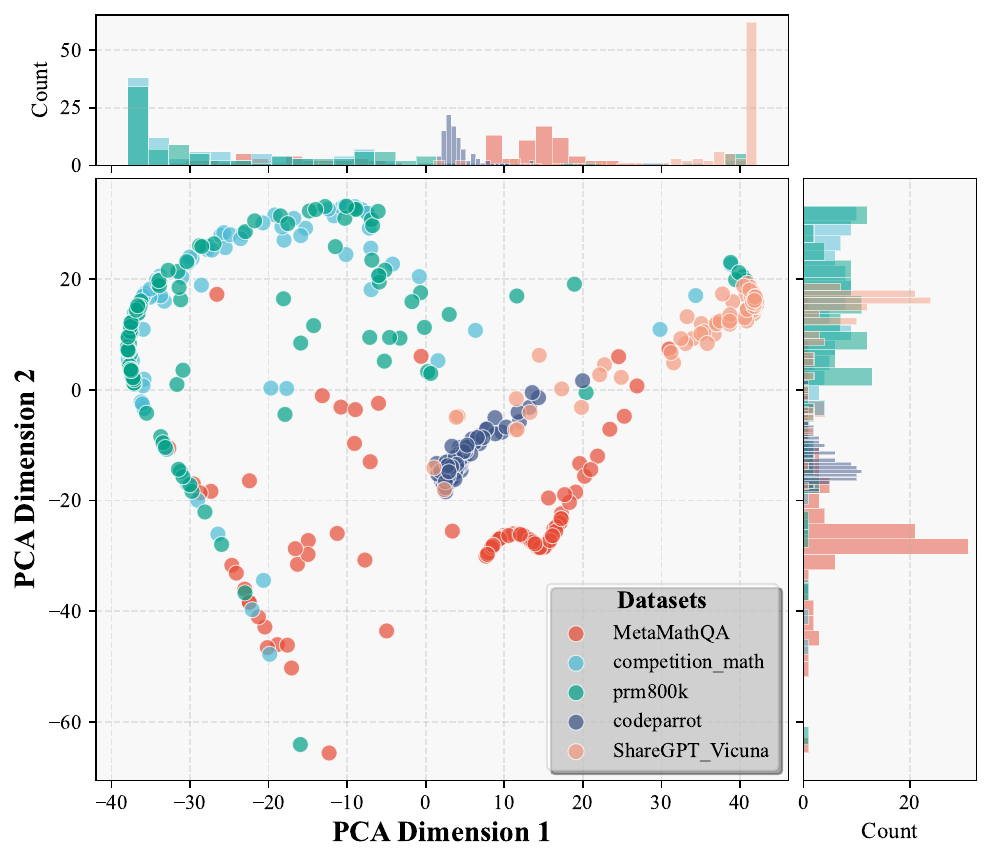}
    \caption{PCA of Generated Prefixes}
    \label{fig:pca-emb1}
  \end{subfigure}
  \begin{subfigure}{0.23\linewidth}
    \centering
    \includegraphics[width=1\linewidth]{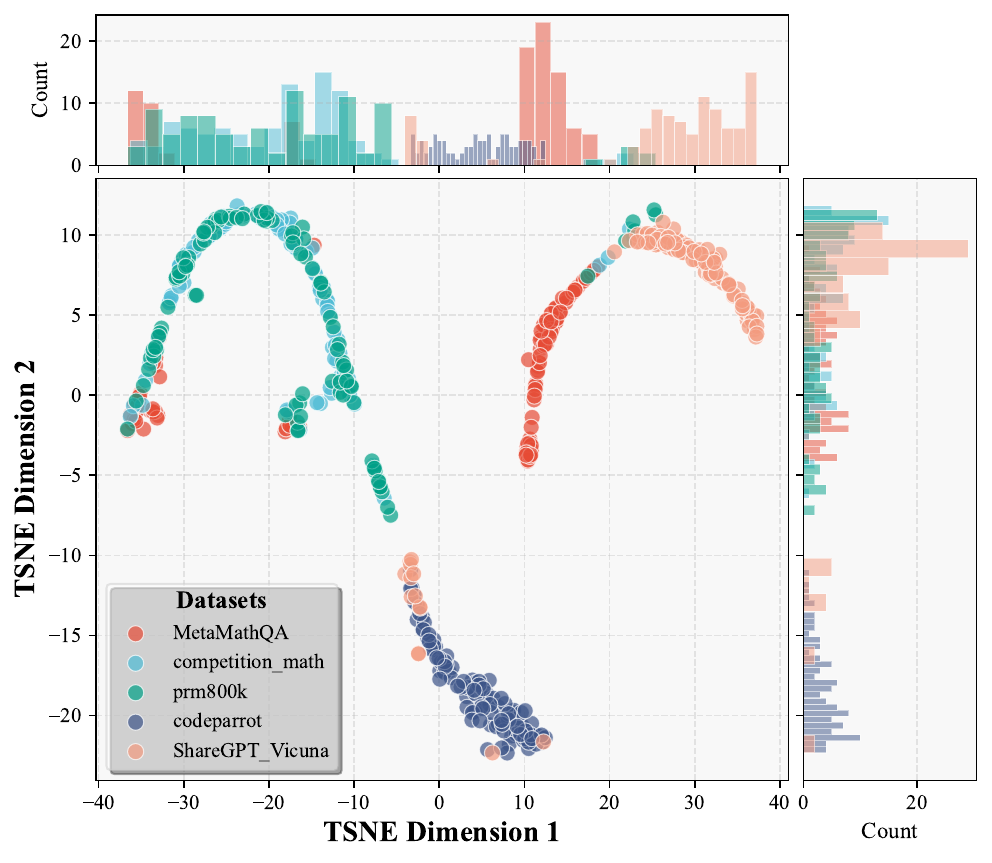}
    \caption{t-SNE of Generated Prefixes}
    \label{fig:tsne-emb1}
  \end{subfigure}
    \begin{subfigure}{0.23\linewidth}
    \centering
    \includegraphics[width=1\linewidth]{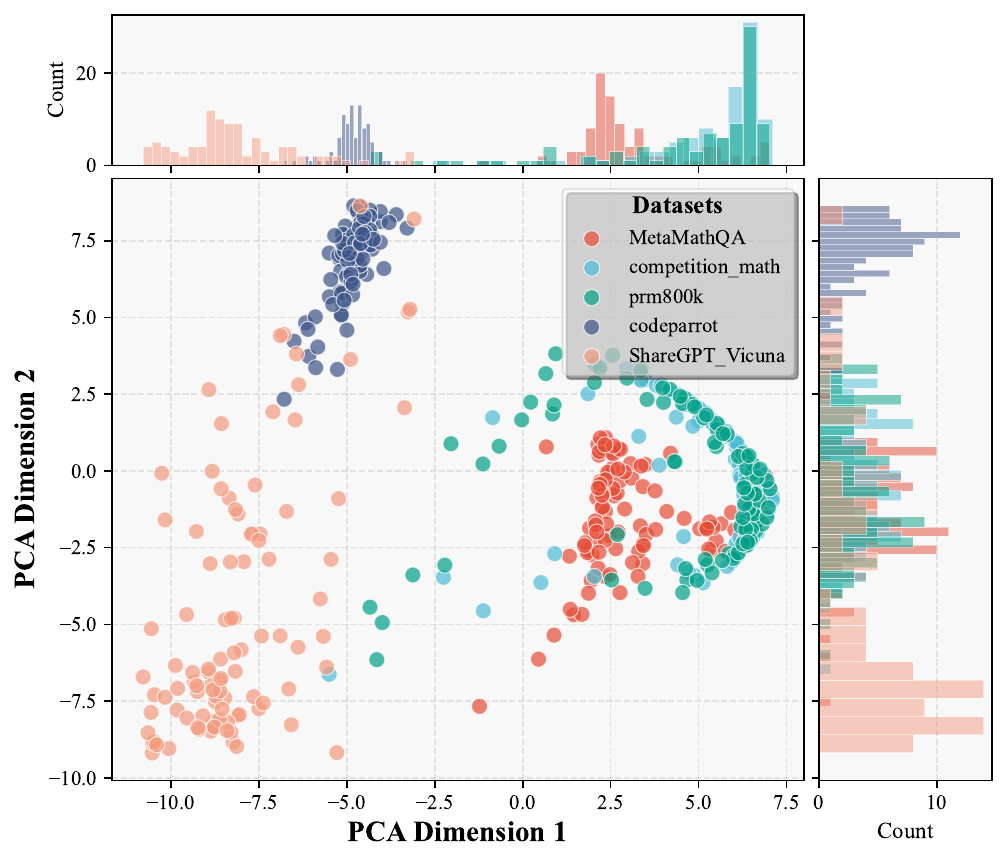}
    \caption{PCA of Latents}
    \label{fig:pca-emb2}
  \end{subfigure}
  \begin{subfigure}{0.23\linewidth}
    \centering
    \includegraphics[width=1\linewidth]{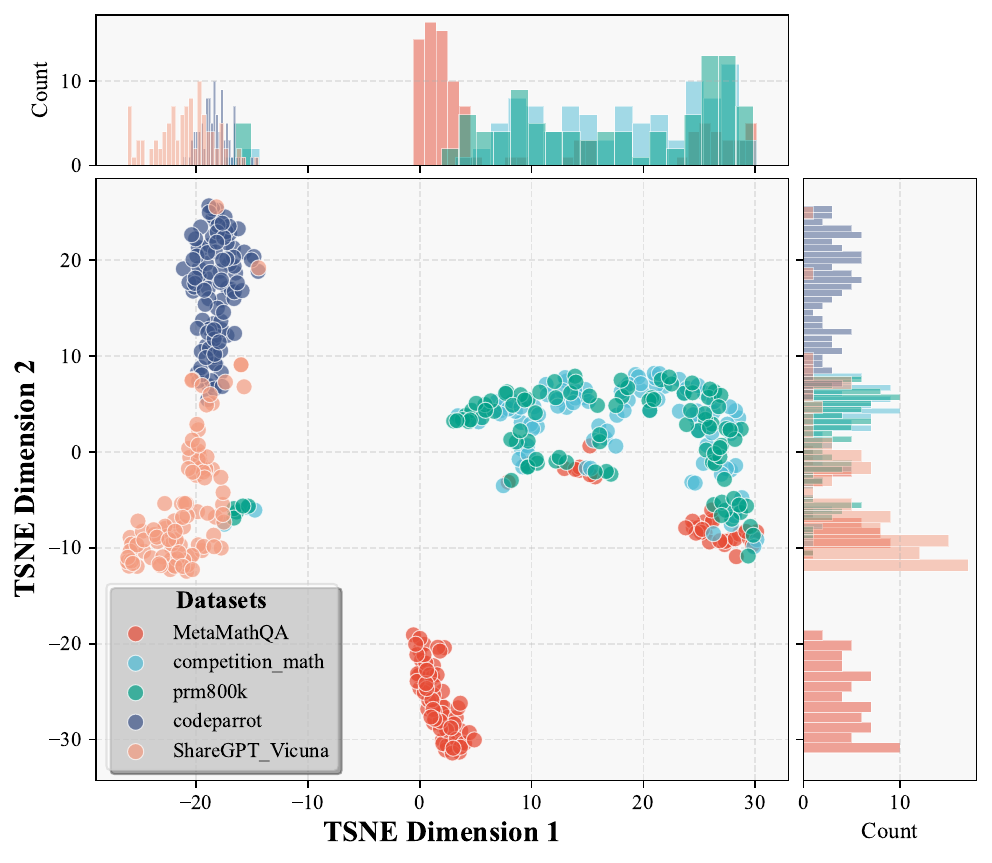}
    \caption{t-SNE of Latents}
    \label{fig:tsne-emb2}
  \end{subfigure}
  \vspace{-5pt}
  \caption{Visualization of the learned latent space and generated prefix embeddings via PCA and t-SNE. Left two panels: projections of decoded prefix embeddings $D_\psi(\mathbf{z})$ (colored by domains). Right two panels: projections of the corresponding latent vectors $\mathbf{z} = E_\phi(\text{mean-pool}(\mathbf{q};\mathbf{o}))$. Clear clustering by reasoning domain in both spaces confirms that the VAE disentangles high-level reasoning strategies into distinct regions of the latent space.}
  \label{fig:dimension-reduction}
  \vspace{-10pt}
\end{figure*}

\subsection{Evaluation of Latent-Guided Inference}

\textbf{Direct Gaussian Noise Injection.} Fig.~\ref{fig:motivation} illustrates a key finding motivating our methodology: injecting a single Gaussian noise token (i.e., sampling $z \sim \mathcal{N}(0,I)$ and using $D_\psi(z)$ as a prefix) before the prompt embeddings, while using greedy decoding for each candidate, yields substantial gains in pass@$k$ accuracy across multiple benchmarks. For instance, on GSM8K, pass@32 improves from 52.9\% to 85.3\% with latent sampling, despite no change in decoding strategy per sample. This demonstrates that performance gains stem not from token-level stochasticity but from strategic variation induced by the latent context. However, naively injecting random noise often degrades performance due to misalignment with the model's native embedding space. To address this, we design our framework to ensure that latent-guided prefixes remain both diversity-influential and well-aligned with the embedding distribution, enabling controllable and effective inference.

\begin{table}[tb!]
\centering
\small
\setlength{\tabcolsep}{4pt}
\caption{Pass@8 results of latent-guided inference.}
\vspace{-10pt}
\label{tab:math_results}
\begin{tabular}{lccc}
\toprule
\textbf{Latent Source} & \textbf{MATH500} & \textbf{Olympic} & \textbf{GSM8K} \\
\midrule
codeparrot (Code) & 70.8 & 42.95 & 94.47 \\
MetaMathQA (Math) & \textbf{72.4} & \textbf{46.11} & \textbf{95.0} \\
ShareGPT\_Vicuna (QA) & 71.0 & 45.47 & 93.63 \\
\bottomrule
\end{tabular}
\vspace{-10pt}
\end{table}

\noindent\textbf{Latent-Guided Inference on LLMs: Targeted Intervention via Biased Latent Sampling.} We evaluate latent-guided inference on a supervised fine-tuned (SFT) Qwen3-4B-Base model. The VAE is trained on a curated dataset of 5K high-quality question–answer pairs spanning mathematics, code generation, and multi-hop question answering. Specifically, the training data includes MetaMathQA~\cite{yu2023metamath}, competition\_math~\cite{hendrycksmath2021}, PRM800K~\cite{lightman2023lets}, ShareGPT\_Vicuna\_unfiltered~\cite{vicuna2023}, and CodeParrot~\cite{tunstall2022natural}. After SFT adaptation ($10$ iterations with prefix length $L=1$), we assess pass@8 performance under latent sampling during inference to demonstrate controllable reasoning. 

We perform domain-specific intervention by leveraging the structure of the latent space. Specifically, we collect representative reasoning trajectories from three distinct domains: mathematical problem solving (e.g., from MetaMathQA), code generation (e.g., from CodeParrot), and general question answering (e.g., from ShareGPT\_Vicuna), and encode them into the latent space using the trained encoder $E_\phi$. For each domain, we compute the empirical mean and covariance of its latent embeddings, thereby identifying well-separated regions that correspond to characteristic reasoning styles. During inference, given a new prompt, we restrict latent sampling to the region associated with the given domain and decode it via $D_\psi(z)$ to obtain a domain-aligned prefix. As shown in Table~\ref{tab:math_results}, using math-domain latents yields the best performance across all mathematical benchmarks: it achieves \textbf{72.4} on MATH500, \textbf{46.11} on Olympic, and \textbf{95.0} on GSM8K, consistently outperforming code- or QA-biased latents.


\noindent\textbf{Latent-Guided Inference on VLMs: Enhancing Grounding via Structured Diversity.} We further extend our framework to vision-language models (VLMs) and evaluate its effectiveness on referring expression comprehension, a canonical grounding task where the model must localize an object in an image based on a natural language description. Following the evaluation protocol in VLM-R1~\cite{shen2025vlm}, we measure accuracy as the fraction of predictions whose bounding box achieves an IoU $\geq 0.5$ with the ground-truth annotation.

Our VLM architecture builds upon a Qwen2.5VL-3B backbone: visual inputs are encoded via a ViT into a sequence of image tokens, which replace special image placeholders in the text prompt. To enable latent-guided control, we sample a noise vector $\mathbf{z} \in \mathbb{R}^{1024}$ from $\mathcal{N}(0, I)$, reshape it into an $8 \times 128$ matrix, and pass it through a lightweight GPT-style decoder to produce $8$ prefix embeddings $\mathbf{A} \in \mathbb{R}^{8 \times d}$. These are prepended to the prompt embedding (which already includes the image tokens) before being fed into the LLM decoder. During inference, we compare four configurations: 1) Baseline (greedy): standard greedy decoding without latent prefixes; 2) Baseline + sampling: stochastic decoding (e.g., temperature=0.7) without latent control; 3) Latent-guided (greedy): greedy decoding with latent prefixes ($L=8$); 4) Latent-guided + sampling: stochastic decoding combined with latent prefixes.

\begin{figure}[tb!]
  \centering
    \includegraphics[width=0.9\linewidth]{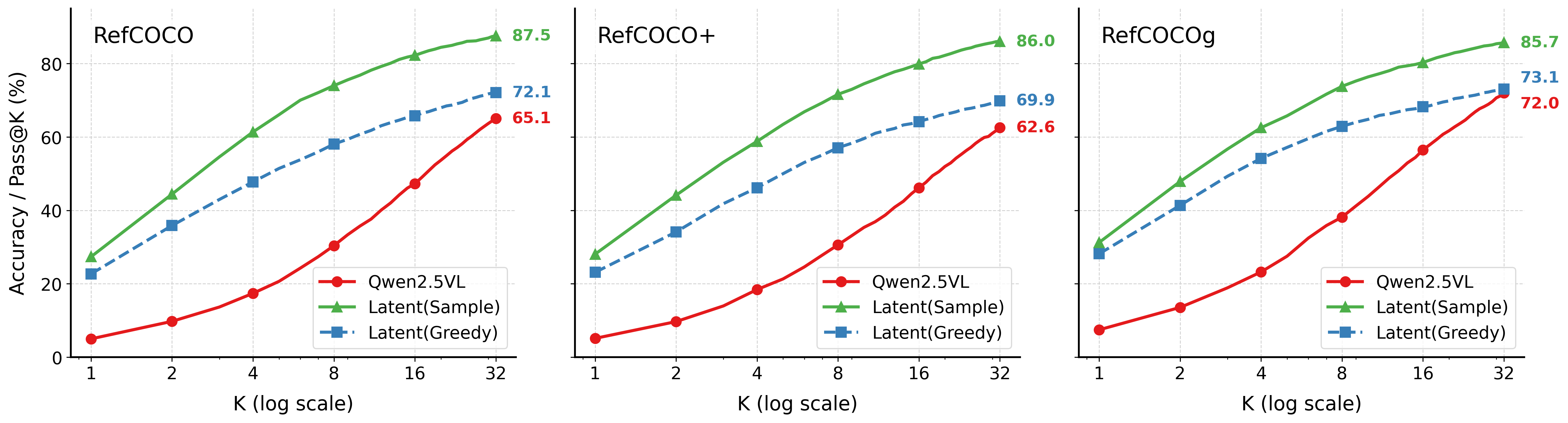}
    \vspace{-10pt}
    \caption{Pass@32 curves on the RefCOCO datasets.}
    \label{fig:refcoco_curves}
    \vspace{-5pt}
\end{figure}

We evaluate all variants using pass@32 on three standard benchmarks: RefCOCO~\cite{yu2016modeling}, RefCOCO+~\cite{yu2016modeling}, and RefCOCOg~\cite{mao2016generation}. As shown in Table~\ref{tab:vlm_results} and Fig.~\ref{fig:refcoco_curves}, latent-guided inference significantly boosts performance under greedy decoding. Notably, \textit{latent-guided + sampling} achieves the best results across all datasets, confirming that latent diversity and decoding stochasticity are complementary. Crucially, the improvement from latent guidance alone (greedy vs. latent-greedy) exceeds that from sampling alone (greedy vs. baseline+sampling), underscoring the unique value of structured, pre-generative exploration in multimodal reasoning. These results validate that Reasoning Palette generalizes beyond pure language tasks: by injecting domain-agnostic yet semantically rich latent context, it enhances the model's ability to explore diverse grounding hypotheses, leading to more robust and accurate spatial reasoning.

\begin{figure}[tb!]
  \centering
  \includegraphics[width=0.9\linewidth]{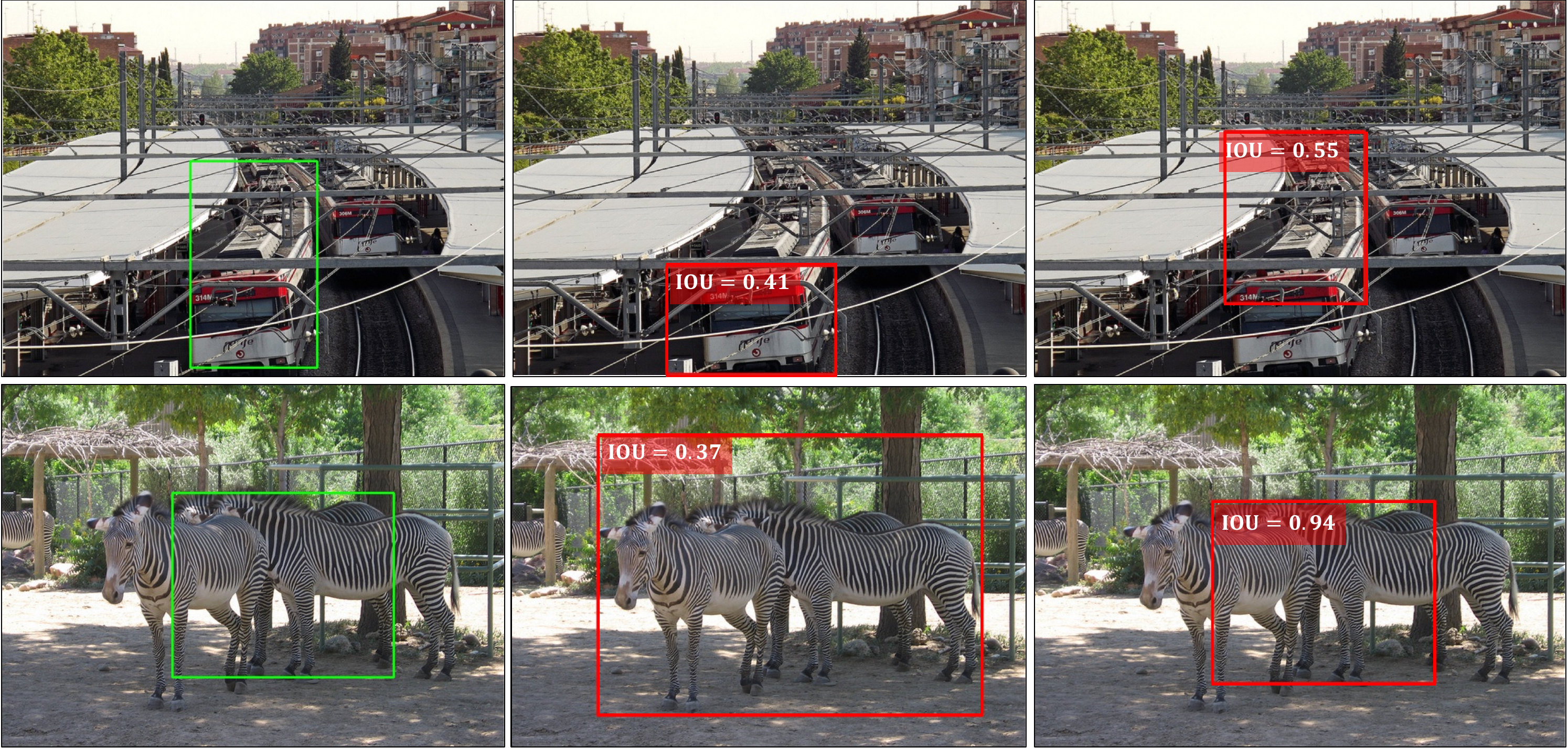} 
  \vspace{-5pt}
  \caption{Qualitative results on RefCOCO dataset. From left to right: input image with the ground-truth bounding box, prediction from Qwen2.5VL-3B (greedy decoding), and prediction from our method, Qwen2.5VL-3B (greedy decoding) with a randomly sampled latent. The referring expressions for the top and bottom rows are \textit{\textbf{train closest to the bottom}} and \textit{\textbf{a zebra standing behind two other zebras, with only its mane and rear showing}}, respectively.}
  \label{fig:refcoco_vis}
  \vspace{-13pt}
\end{figure}

Upon analysis, we find that under the Qwen2.5VL-3B with greedy decoding setting, the model often correctly identifies the target entity but fails to produce outputs in the required format, leading to low metric scores. The performance improves substantially when evaluated with multiple sampling passes (Pass@32). Fig.~\ref{fig:refcoco_vis} shows qualitative results, where introducing the latent leads to significant performance improvements on challenging cases.

\begin{table}[tb!]
\centering
\small
\setlength{\tabcolsep}{6pt}
\caption{Pass@32 on referring expression comprehension.}
\vspace{-10pt}
\label{tab:vlm_results}
\resizebox{\linewidth}{!}{
\begin{tabular}{lccc}
\toprule
\textbf{Method} & \textbf{RefCOCO} & \textbf{RefCOCO+} & \textbf{RefCOCOg} \\
\midrule
Baseline (greedy) & 2.0 & 2.0 & 4.67 \\
Baseline + sampling & 65.07 & 62.57 & 72.0 \\
Latent-guided (greedy) & 72.07 & 73.07 & 73.1 \\
Latent-guided + sampling & \textbf{87.53} & \textbf{86.03} & \textbf{85.7} \\
\bottomrule
\end{tabular}}
\vspace{-10pt}
\end{table}

\noindent\textbf{Latent Space Analysis.} To verify the controllability of our latent space, we perform post-hoc analysis using the same dataset employed for VAE training. We collect 500 reasoning trajectories covering the five datasets, including MetaMathQA (Math), competition\_math (Math), PRM800K (Math), ShareGPT\_Vicuna\_unfiltered (QA), and CodeParrot (Code), 100 trajectories for each. Each $(\mathbf{q},\mathbf{o})$ pair is encoded into its latent representation $\mathbf{z} = E_\phi(\text{mean-pool}(\mathbf{q};\mathbf{o}))$, and we generate prefix tokens using the latents $\mathbf{p}=D_\phi(\mathbf{z})$. We visualize the resulting latents and the corresponding prefixes using dimensionality reduction techniques including PCA~\cite{abdi2010principal} and t-SNE~\cite{maaten2008visualizing}. 

The visualization in Fig.~\ref{fig:dimension-reduction} reveals interpretable clustering aligned with reasoning domains. Notably, the two advanced Math datasets {competition\_math} and {PRM800K} exhibit highly overlapping latent distributions, reflecting their shared focus on rigorous, formal mathematical reasoning. In contrast, {MetaMathQA}, while still within the mathematical cluster, occupies a slightly offset region, likely due to its emphasis on step-by-step pedagogical explanations rather than competition-level deductions. This subtle separation demonstrates the latent space's sensitivity to fine-grained reasoning styles even within the same broad domain.

Meanwhile, code generation (CodeParrot) and general QA (ShareGPT\_Vicuna) form two more separated clusters from the mathematical region, underscoring the divergence in their underlying reasoning patterns. Interestingly, the QA cluster displays the largest spatial spread, with some outlier samples extending toward the code and math regions. This suggests that open-ended question answering encompasses a broader spectrum of reasoning strategies, including occasional algorithmic or quantitative elements, consistent with its heterogeneous nature. Crucially, this structural coherence is preserved not only in the latent vectors $z$ but also in their decoded prefix tokens $\mathbf{p} = D_\psi(\mathbf{z})$. This mutual consistency confirms that the VAE has successfully learned a disentangled and functional mapping, one that captures high-level reasoning semantics and translates them into effective behavioral guidance for the model.


\subsection{Reinforcement Learning with Latent Guidance}

\begin{table*}[!tb]
\makeatletter\def\@captype{table}\makeatother
\caption{Results of Math reasoning over different model sizes. Bold denotes the best results.}
\vspace{-10pt}
\centering
\resizebox{0.9\linewidth}{!}{
\begin{tabular}{lccccccc}
\toprule
\textbf{Method} & \textbf{MATH500} & \textbf{OlympiadBench} & \textbf{AMC23} & \textbf{GSM8K}  & \textbf{MinervaMath} & \textbf{Avg.}
\\
\midrule
\multicolumn{7}{c}{\textit{Qwen3-1.7B-Base}} \\
\midrule
GRPO & 23.35 & 8.05 & 10.31 & 51.86  & 12.36 &   21.18
\\
\rowcolor{blue!7} + Reasoning Palette (Two-Phase) & 26.02 \blueplus{+2.67} & 9.63 \blueplus{+1.58} & 9.69 \loss{-0.62} & \textbf{57.23} \blueplus{+5.37}  & 11.99 \loss{-0.37} & 22.91 \blueplus{+1.73} 
\\
\rowcolor{blue!7} + Reasoning Palette (Linear Decay) & \textbf{27.32} \blueplus{+3.97} & \textbf{9.89} \blueplus{+1.84} & \textbf{12.80} \blueplus{+2.49} & 57.00 \blueplus{+5.14}  & \textbf{12.82} \blueplus{+0.46} & \textbf{24.05} \blueplus{+2.87} 
\\
RLOO & 24.75 & 8.61 & 7.50 & 54.96  & 13.05 & 21.77
\\
\rowcolor{blue!7} + Reasoning Palette (Two-Phase) & \textbf{25.77} \blueplus{+1.02} & \textbf{9.50} \blueplus{+0.89} & 10.31 \blueplus{+2.81} & 55.83 \blueplus{+0.87}  & 12.59 \loss{-0.46} & 22.80 \blueplus{+1.03} 
\\
\rowcolor{blue!7} + Reasoning Palette (Linear Decay) & 25.72 \blueplus{+0.97} & 8.87 \blueplus{+0.26} & \textbf{11.25} \blueplus{+3.75} & \textbf{57.62} \blueplus{+2.66}  & \textbf{13.98} \blueplus{+0.93} & \textbf{23.49} \blueplus{+1.72} 
\\
\midrule
\multicolumn{7}{c}{\textit{Qwen3-4B-Base}} \\
\midrule
GRPO & 68.65 & 41.32 & 50.94 & 91.05  & 39.39 &   58.27
\\
\rowcolor{blue!7} + Reasoning Palette (Two-Phase) & 70.53 \blueplus{+1.88} & 43.95 \blueplus{+2.63} & \textbf{55.00} \blueplus{+4.06} & 92.19 \blueplus{+1.14}  & 42.20 \blueplus{+2.81} & \textbf{60.77} \blueplus{+2.50} 
\\
\rowcolor{blue!7} + Reasoning Palette (Linear Decay) & \textbf{72.67} \blueplus{+4.02} & \textbf{45.29} \blueplus{+3.97} & 47.50 \loss{-3.44} & \textbf{92.64} \blueplus{+1.59}  & \textbf{42.53} \blueplus{+3.14} & 60.12 \blueplus{+1.85} 
\\
RLOO & 74.55 & 43.53 & 51.25 & 91.55  & 36.07 & 59.39
\\
\rowcolor{blue!7} + Reasoning Palette (Two-Phase) & \textbf{75.90} \blueplus{+1.35} & 44.21 \blueplus{+0.68} & 52.81 \blueplus{+1.56} & 91.80 \blueplus{+0.25}  & 38.75 \blueplus{+2.68} & 60.69 \blueplus{+1.30} 
\\
\rowcolor{blue!7} + Reasoning Palette (Linear Decay) & 71.50 \loss{-3.05} & \textbf{44.76} \blueplus{+1.23} & \textbf{55.62} \blueplus{+4.37} & \textbf{91.85} \blueplus{+0.3}  & \textbf{42.67} \blueplus{+6.6} & \textbf{61.28} \blueplus{+1.89} 
\\

\midrule
\multicolumn{7}{c}{\textit{Qwen3-8B-Base}} \\
\midrule
GRPO & 70.05 & 44.34 & 54.69 & 92.13  & 39.07 &   60.05
\\
\rowcolor{blue!7} + Reasoning Palette (Two-Phase) & 70.38 \blueplus{+0.33} & 43.24 \loss{-1.10} & 55.62 \blueplus{+0.93} & \textbf{92.24} \blueplus{+0.11}  & \textbf{41.37} \blueplus{+2.30} & 60.57 \blueplus{+0.52} 
\\
\rowcolor{blue!7} + Reasoning Palette (Linear Decay) & \textbf{70.78} \blueplus{+0.73} & \textbf{44.74} \blueplus{+0.40} & \textbf{57.19} \blueplus{+2.50} & 92.22 \blueplus{+0.09}  & 40.91 \blueplus{+1.84} & \textbf{61.17} \blueplus{+1.12} 
\\
RLOO & 69.53 & 43.76 & 55.00 & 91.82  & 39.48 & 59.91
\\
\rowcolor{blue!7} + Reasoning Palette (Two-Phase) & \textbf{72.27} \blueplus{+2.74} & 46.00 \blueplus{+2.24} & 57.50 \blueplus{+2.50} & \textbf{93.04} \blueplus{+1.22}  & 42.57 \blueplus{+3.09} & 62.28 \blueplus{+2.37} 
\\
\rowcolor{blue!7} + Reasoning Palette (Linear Decay) & 72.20 \blueplus{+2.67} & \textbf{46.61} \blueplus{+2.85} & \textbf{59.38} \blueplus{+4.38} & 93.03 \blueplus{+1.21}  & \textbf{43.77} \blueplus{+4.29} & \textbf{63.00} \blueplus{+3.09} 
\\

\bottomrule
\end{tabular}
}
\vspace{-5pt}
\label{tab:math_reasoning}
\end{table*}

We now present a comprehensive evaluation of Reasoning Palette within reinforcement learning (RL) training, focusing on its ability to enhance exploration and final performance on complex reasoning tasks. Our experiments are built upon two state-of-the-art RL algorithms for language models: Group Relative Policy Optimization (GRPO)~\citep{shao2024deepseekmath} and Reward-Label Optimized Off-policy (RLOO)~\citep{guo2024directlanguagemodelalignment}. Both methods rely on verifiable, outcome-based rewards

All models are initialized from the SFT-adapted checkpoints as described in Sec.~\ref{subsec:sft}, ensuring sensitivity to latent-conditioned prefixes while preserving general capabilities. The baselines are finetuned with standard SFT for fair comparison. Training is conducted on the \textit{DeepMath} dataset~\citep{he2025deepmath}, a large-scale collection of mathematical problems with verified solutions and step-by-step rationales.

We compare two latent scheduling strategies that modulate the strength of reasoning-mode control during RL, i.e., \textbf{Two-phase} and \textbf{Linear decay}, as detailed in Sec.~\ref{subsec:rl}. All models generate 8 responses per prompt during rollouts for advantage estimation. Evaluation uses pass@1 for mathematical datasets including AMC23~\cite{li2024numinamath}, GSM8K~\cite{cobbe2021gsm8k}, MinervaMath~\cite{lewkowycz2022solving}, MATH500~\cite{hendrycks2021measuring}, and OlympiadBench~\cite{he2024olympiadbench}.


\noindent\textbf{Main Results.} Table~\ref{tab:math_reasoning} reports performance across five core mathematical reasoning benchmarks. We evaluate on both Qwen3-1.7B-Base, Qwen3-4B-Base, and Qwen3-8B-Base backbones to assess scalability. Across all settings, Reasoning Palette consistently outperforms the respective RL baselines, demonstrating the robustness of our approach. On Qwen3-8B-Base + RLOO, Reasoning Palette improves average performance by +3.09 points, with the largest gains on complex domains (AMC23: +4.38, MinervaMath: +4.29). The linear-decay schedule yields slightly better final performance than the two-phase schedule (+0.75 points on average), suggesting that a smooth transition enables the model to adapt more gracefully during training, facilitating a more effective shift from exploration to exploitation and ultimately leading to higher asymptotic performance.

\begin{figure}[tb!]
  \centering
    \includegraphics[width=0.7\linewidth]{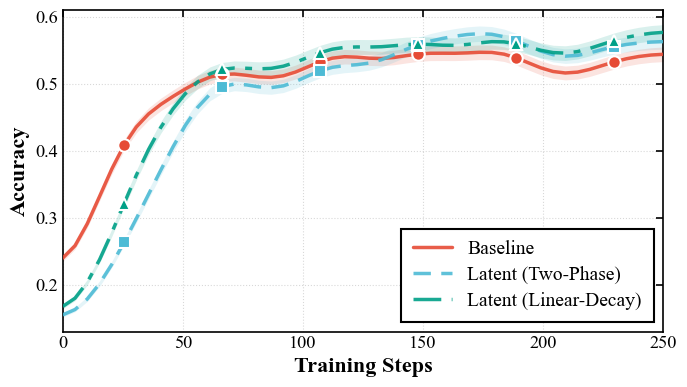}
    \vspace{-10pt}
    \caption{Performance curves of the GRPO baseline and the proposed latent variants during training. Latent variants perform more thorough exploration in early stages of training and then shift toward exploitation in the latter stages, resulting in a performance curve that gradually overtakes baselines.}
    \label{fig:curves}
    \vspace{-15pt}
\end{figure}

\noindent\textbf{Learning Dynamics and Analysis.} Fig.~\ref{fig:curves} shows the training curves of the GRPO baseline and the proposed latent variants, indicating that the gain from injecting a latent exploration signal is not a simple one-shot boost but a change in the whole exploration-exploitation trajectory. The latent variants explore more broadly in the early phase of training which produces slower initial gains in pointwise accuracy but exposes the learner to higher‑quality regions of policy space. As training progresses, the latent diversity signal is reduced, and the policy concentrates on the better behaviors discovered earlier. The net effect is a smoother transition from exploration to exploitation and a modest but consistent improvement in final accuracy relative to the GRPO baseline. Here we compare two latent scheduling strategies. The two-phase control strategy exhibits a clear performance improvement in the latter half stage, whereas the linear-decay strategy transitions gradually from exploration to exploitation, resulting in a smoother overall transition.




\vspace{-10pt}
\section{Conclusion}
\label{sec:conclusion}

We introduce \textbf{Reasoning Palette}, a lightweight framework that shifts exploration in LLM reinforcement learning from token-level randomness to strategy-level diversity by sampling from a VAE-learned latent space of reasoning patterns. These latents are decoded into prefix embeddings that steer the model's internal planning before generation. With a brief SFT warm-up and a two-phase RL schedule (exploration with latent prefixes followed by exploitation without), our method consistently boosts performance across reasoning benchmarks and enables interpretable, domain-aware control during inference, demonstrating that structured latent modulation is a powerful and practical approach to enhancing reasoning in large models.

{
    \small
    \bibliographystyle{ieeenat_fullname}
    \bibliography{main}

@article{ouyang2022training,
  title={Training language models to follow instructions with human feedback},
  author={Ouyang, Long and Wu, Jeffrey and Jiang, Xu and Almeida, Diogo and Wainwright, Carroll and Mishkin, Pamela and Zhang, Chong and Agarwal, Sandhini and Slama, Katarina and Ray, Alex and others},
  journal={Advances in neural information processing systems},
  volume={35},
  pages={27730--27744},
  year={2022}
}

@article{schulman2017proximal,
  title={Proximal policy optimization algorithms},
  author={Schulman, John and Wolski, Filip and Dhariwal, Prafulla and Radford, Alec and Klimov, Oleg},
  journal={arXiv preprint arXiv:1707.06347},
  year={2017}
}

@article{shao2024deepseekmath,
  title={Deepseekmath: Pushing the limits of mathematical reasoning in open language models},
  author={Shao, Zhihong and Wang, Peiyi and Zhu, Qihao and Xu, Runxin and Song, Junxiao and Bi, Xiao and Zhang, Haowei and Zhang, Mingchuan and Li, YK and Wu, Yang and others},
  journal={arXiv preprint arXiv:2402.03300},
  year={2024}
}

@article{williams1992simple,
  title={Simple statistical gradient-following algorithms for connectionist reinforcement learning},
  author={Williams, Ronald J},
  journal={Machine learning},
  volume={8},
  number={3},
  pages={229--256},
  year={1992},
  publisher={Springer}
}

@article{rafailov2023direct,
  title={Direct preference optimization: Your language model is secretly a reward model},
  author={Rafailov, Rafael and Sharma, Archit and Mitchell, Eric and Manning, Christopher D and Ermon, Stefano and Finn, Chelsea},
  journal={Advances in neural information processing systems},
  volume={36},
  pages={53728--53741},
  year={2023}
}

@article{meng2024simpo,
  title={Simpo: Simple preference optimization with a reference-free reward},
  author={Meng, Yu and Xia, Mengzhou and Chen, Danqi},
  journal={Advances in Neural Information Processing Systems},
  volume={37},
  pages={124198--124235},
  year={2024}
}

@article{ethayarajh2024kto,
  title={Kto: Model alignment as prospect theoretic optimization},
  author={Ethayarajh, Kawin and Xu, Winnie and Muennighoff, Niklas and Jurafsky, Dan and Kiela, Douwe},
  journal={arXiv preprint arXiv:2402.01306},
  year={2024}
}

@article{lambert2024tulu,
  title={Tulu 3: Pushing frontiers in open language model post-training},
  author={Lambert, Nathan and Morrison, Jacob and Pyatkin, Valentina and Huang, Shengyi and Ivison, Hamish and Brahman, Faeze and Miranda, Lester James V and Liu, Alisa and Dziri, Nouha and Lyu, Shane and others},
  journal={arXiv preprint arXiv:2411.15124},
  year={2024}
}

@article{guo2025deepseek,
  title={Deepseek-r1: Incentivizing reasoning capability in llms via reinforcement learning},
  author={Guo, Daya and Yang, Dejian and Zhang, Haowei and Song, Junxiao and Zhang, Ruoyu and Xu, Runxin and Zhu, Qihao and Ma, Shirong and Wang, Peiyi and Bi, Xiao and others},
  journal={arXiv preprint arXiv:2501.12948},
  year={2025}
}

@misc{openai2024learning,
author = {OpenAI},
title = {Learning to Reason with LLMs},
year = {2024},
url = {https://openai.com/index/learning-to-reason-with-llms/},
urldate = {2025-05-01}
}

@article{team2025kimi,
  title={Kimi k1. 5: Scaling reinforcement learning with llms},
  author={Team, Kimi and Du, Angang and Gao, Bofei and Xing, Bowei and Jiang, Changjiu and Chen, Cheng and Li, Cheng and Xiao, Chenjun and Du, Chenzhuang and Liao, Chonghua and others},
  journal={arXiv preprint arXiv:2501.12599},
  year={2025}
}

@article{yang2025qwen3,
  title={Qwen3 technical report},
  author={Yang, An and Li, Anfeng and Yang, Baosong and Zhang, Beichen and Hui, Binyuan and Zheng, Bo and Yu, Bowen and Gao, Chang and Huang, Chengen and Lv, Chenxu and others},
  journal={arXiv preprint arXiv:2505.09388},
  year={2025}
}

@article{yu2025dapo,
  title={Dapo: An open-source llm reinforcement learning system at scale},
  author={Yu, Qiying and Zhang, Zheng and Zhu, Ruofei and Yuan, Yufeng and Zuo, Xiaochen and Yue, Yu and Dai, Weinan and Fan, Tiantian and Liu, Gaohong and Liu, Lingjun and others},
  journal={arXiv preprint arXiv:2503.14476},
  year={2025}
}

@article{yue2025vapo,
  title={Vapo: Efficient and reliable reinforcement learning for advanced reasoning tasks},
  author={Yue, Yu and Yuan, Yufeng and Yu, Qiying and Zuo, Xiaochen and Zhu, Ruofei and Xu, Wenyuan and Chen, Jiaze and Wang, Chengyi and Fan, TianTian and Du, Zhengyin and others},
  journal={arXiv preprint arXiv:2504.05118},
  year={2025}
}

@article{wang2025beyond,
  title={Beyond the 80/20 rule: High-entropy minority tokens drive effective reinforcement learning for llm reasoning},
  author={Wang, Shenzhi and Yu, Le and Gao, Chang and Zheng, Chujie and Liu, Shixuan and Lu, Rui and Dang, Kai and Chen, Xionghui and Yang, Jianxin and Zhang, Zhenru and others},
  journal={arXiv preprint arXiv:2506.01939},
  year={2025}
}

@article{cobbe2021training,
  title={Training verifiers to solve math word problems},
  author={Cobbe, Karl and Kosaraju, Vineet and Bavarian, Mohammad and Chen, Mark and Jun, Heewoo and Kaiser, Lukasz and Plappert, Matthias and Tworek, Jerry and Hilton, Jacob and Nakano, Reiichiro and others},
  journal={arXiv preprint arXiv:2110.14168},
  year={2021}
}

@article{li2024numinamath,
  title={Numinamath: The largest public dataset in ai4maths with 860k pairs of competition math problems and solutions},
  author={Li, Jia and Beeching, Edward and Tunstall, Lewis and Lipkin, Ben and Soletskyi, Roman and Huang, Shengyi and Rasul, Kashif and Yu, Longhui and Jiang, Albert Q and Shen, Ziju and others},
  journal={Hugging Face repository},
  volume={13},
  number={9},
  pages={9},
  year={2024}
}

@article{chen2021evaluating,
  title={Evaluating large language models trained on code},
  author={Chen, Mark and Tworek, Jerry and Jun, Heewoo and Yuan, Qiming and Pinto, Henrique Ponde De Oliveira and Kaplan, Jared and Edwards, Harri and Burda, Yuri and Joseph, Nicholas and Brockman, Greg and others},
  journal={arXiv preprint arXiv:2107.03374},
  year={2021}
}

@article{yang2024qwen2,
  title={Qwen2. 5-math technical report: Toward mathematical expert model via self-improvement},
  author={Yang, An and Zhang, Beichen and Hui, Binyuan and Gao, Bofei and Yu, Bowen and Li, Chengpeng and Liu, Dayiheng and Tu, Jianhong and Zhou, Jingren and Lin, Junyang and others},
  journal={arXiv preprint arXiv:2409.12122},
  year={2024}
}

@article{liu2023agentbench,
  title={Agentbench: Evaluating llms as agents},
  author={Liu, Xiao and Yu, Hao and Zhang, Hanchen and Xu, Yifan and Lei, Xuanyu and Lai, Hanyu and Gu, Yu and Ding, Hangliang and Men, Kaiwen and Yang, Kejuan and others},
  journal={arXiv preprint arXiv:2308.03688},
  year={2023}
}

@article{yao2024tau,
  title={$\tau$-bench: A Benchmark for Tool-Agent-User Interaction in Real-World Domains},
  author={Yao, Shunyu and Shinn, Noah and Razavi, Pedram and Narasimhan, Karthik},
  journal={arXiv preprint arXiv:2406.12045},
  year={2024}
}

@article{hui2024qwen2,
  title={Qwen2. 5-coder technical report},
  author={Hui, Binyuan and Yang, Jian and Cui, Zeyu and Yang, Jiaxi and Liu, Dayiheng and Zhang, Lei and Liu, Tianyu and Zhang, Jiajun and Yu, Bowen and Lu, Keming and others},
  journal={arXiv preprint arXiv:2409.12186},
  year={2024}
}

@article{jimenez2023swe,
  title={Swe-bench: Can language models resolve real-world github issues?},
  author={Jimenez, Carlos E and Yang, John and Wettig, Alexander and Yao, Shunyu and Pei, Kexin and Press, Ofir and Narasimhan, Karthik},
  journal={arXiv preprint arXiv:2310.06770},
  year={2023}
}

@article{hendrycks2021measuring,
  title={Measuring mathematical problem solving with the math dataset},
  author={Hendrycks, Dan and Burns, Collin and Kadavath, Saurav and Arora, Akul and Basart, Steven and Tang, Eric and Song, Dawn and Steinhardt, Jacob},
  journal={arXiv preprint arXiv:2103.03874},
  year={2021}
}

@article{liu2025part,
  title={Part I: Tricks or Traps? A Deep Dive into RL for LLM Reasoning},
  author={Liu, Zihe and Liu, Jiashun and He, Yancheng and Wang, Weixun and Liu, Jiaheng and Pan, Ling and Hu, Xinyu and Xiong, Shaopan and Huang, Ju and Hu, Jian and others},
  journal={arXiv preprint arXiv:2508.08221},
  year={2025}
}

@article{schulman2015high,
  title={High-dimensional continuous control using generalized advantage estimation},
  author={Schulman, John and Moritz, Philipp and Levine, Sergey and Jordan, Michael and Abbeel, Pieter},
  journal={arXiv preprint arXiv:1506.02438},
  year={2015}
}

@article{lewkowycz2022solving,
  title={Solving quantitative reasoning problems with language models},
  author={Lewkowycz, Aitor and Andreassen, Anders and Dohan, David and Dyer, Ethan and Michalewski, Henryk and Ramasesh, Vinay and Slone, Ambrose and Anil, Cem and Schlag, Imanol and Gutman-Solo, Theo and others},
  journal={Advances in neural information processing systems},
  volume={35},
  pages={3843--3857},
  year={2022}
}

@article{hendrycksmath2021,
    title={Measuring Mathematical Problem Solving With the MATH Dataset},
    author={Dan Hendrycks
    and Collin Burns
    and Saurav Kadavath
    and Akul Arora
    and Steven Basart
    and Eric Tang
    and Dawn Song
    and Jacob Steinhardt},
    journal={arXiv preprint arXiv:2103.03874},
    year={2021}
}

@article{lightman2023lets,
      title={Let's Verify Step by Step}, 
      author={Lightman, Hunter and Kosaraju, Vineet and Burda, Yura and Edwards, Harri and Baker, Bowen and Lee, Teddy and Leike, Jan and Schulman, John and Sutskever, Ilya and Cobbe, Karl},
      journal={arXiv preprint arXiv:2305.20050},
      year={2023}
}

@misc{vicuna2023,
    title = {Vicuna: An Open-Source Chatbot Impressing GPT-4 with 90\%* ChatGPT Quality},
    url = {https://lmsys.org/blog/2023-03-30-vicuna/},
    author = {Chiang, Wei-Lin and Li, Zhuohan and Lin, Zi and Sheng, Ying and Wu, Zhanghao and Zhang, Hao and Zheng, Lianmin and Zhuang, Siyuan and Zhuang, Yonghao and Gonzalez, Joseph E. and Stoica, Ion and Xing, Eric P.},
    month = {March},
    year = {2023}
}

@article{abdi2010principal,
  title={Principal component analysis},
  author={Abdi, Herv{\'e} and Williams, Lynne J},
  journal={Wiley interdisciplinary reviews: computational statistics},
  volume={2},
  number={4},
  pages={433--459},
  year={2010},
  publisher={Wiley Online Library}
}

@inproceedings{yu2016modeling,
  title={Modeling context in referring expressions},
  author={Yu, Licheng and Poirson, Patrick and Yang, Shan and Berg, Alexander C and Berg, Tamara L},
  booktitle={European conference on computer vision},
  pages={69--85},
  year={2016},
  organization={Springer}
}

@inproceedings{haarnoja2018soft,
  title={Soft actor-critic: Off-policy maximum entropy deep reinforcement learning with a stochastic actor},
  author={Haarnoja, Tuomas and Zhou, Aurick and Abbeel, Pieter and Levine, Sergey},
  booktitle={International conference on machine learning},
  pages={1861--1870},
  year={2018},
  organization={Pmlr}
}

@inproceedings{mao2016generation,
  title={Generation and comprehension of unambiguous object descriptions},
  author={Mao, Junhua and Huang, Jonathan and Toshev, Alexander and Camburu, Oana and Yuille, Alan L and Murphy, Kevin},
  booktitle={Proceedings of the IEEE conference on computer vision and pattern recognition},
  pages={11--20},
  year={2016}
}

@article{shen2025vlm,
  title={Vlm-r1: A stable and generalizable r1-style large vision-language model},
  author={Shen, Haozhan and Liu, Peng and Li, Jingcheng and Fang, Chunxin and Ma, Yibo and Liao, Jiajia and Shen, Qiaoli and Zhang, Zilun and Zhao, Kangjia and Zhang, Qianqian and others},
  journal={arXiv preprint arXiv:2504.07615},
  year={2025}
}

@inproceedings{huang2023towards,
  title={Towards reasoning in large language models: A survey},
  author={Huang, Jie and Chang, Kevin Chen-Chuan},
  booktitle={Findings of the association for computational linguistics: ACL 2023},
  pages={1049--1065},
  year={2023}
}

@article{zhao2023survey,
  title={A survey of large language models},
  author={Zhao, Wayne Xin and Zhou, Kun and Li, Junyi and Tang, Tianyi and Wang, Xiaolei and Hou, Yupeng and Min, Yingqian and Zhang, Beichen and Zhang, Junjie and Dong, Zican and others},
  journal={arXiv preprint arXiv:2303.18223},
  volume={1},
  number={2},
  year={2023}
}

@article{zhang2024vision,
  title={Vision-language models for vision tasks: A survey},
  author={Zhang, Jingyi and Huang, Jiaxing and Jin, Sheng and Lu, Shijian},
  journal={IEEE transactions on pattern analysis and machine intelligence},
  volume={46},
  number={8},
  pages={5625--5644},
  year={2024},
  publisher={IEEE}
}

@article{ghosh2024exploring,
  title={Exploring the frontier of vision-language models: A survey of current methodologies and future directions},
  author={Ghosh, Akash and Acharya, Arkadeep and Saha, Sriparna and Jain, Vinija and Chadha, Aman},
  journal={arXiv preprint arXiv:2404.07214},
  year={2024}
}

@article{maaten2008visualizing,
  title={Visualizing data using t-SNE},
  author={Maaten, Laurens van der and Hinton, Geoffrey},
  journal={Journal of machine learning research},
  volume={9},
  number={Nov},
  pages={2579--2605},
  year={2008}
}

@book{tunstall2022natural,
  title={Natural language processing with transformers},
  author={Tunstall, Lewis and Von Werra, Leandro and Wolf, Thomas},
  year={2022},
  publisher={" O'Reilly Media, Inc."}
}

@article{yu2023metamath,
  title={MetaMath: Bootstrap Your Own Mathematical Questions for Large Language Models},
  author={Yu, Longhui and Jiang, Weisen and Shi, Han and Yu, Jincheng and Liu, Zhengying and Zhang, Yu and Kwok, James T and Li, Zhenguo and Weller, Adrian and Liu, Weiyang},
  journal={arXiv preprint arXiv:2309.12284},
  year={2023}
}

@article{he2024olympiadbench,
  title={Olympiadbench: A challenging benchmark for promoting agi with olympiad-level bilingual multimodal scientific problems},
  author={He, Chaoqun and Luo, Renjie and Bai, Yuzhuo and Hu, Shengding and Thai, Zhen Leng and Shen, Junhao and Hu, Jinyi and Han, Xu and Huang, Yujie and Zhang, Yuxiang and others},
  journal={arXiv preprint arXiv:2402.14008},
  year={2024}
}

@article{ziegler2019fine,
  title={Fine-tuning language models from human preferences},
  author={Ziegler, Daniel M and Stiennon, Nisan and Wu, Jeffrey and Brown, Tom B and Radford, Alec and Amodei, Dario and Christiano, Paul and Irving, Geoffrey},
  journal={arXiv preprint arXiv:1909.08593},
  year={2019}
}

@article{christiano2017deep,
  title={Deep reinforcement learning from human preferences},
  author={Christiano, Paul F and Leike, Jan and Brown, Tom and Martic, Miljan and Legg, Shane and Amodei, Dario},
  journal={Advances in neural information processing systems},
  volume={30},
  year={2017}
}

@article{schulman2022chatgpt,
  title={Chatgpt: Optimizing language models for dialogue},
  author={Schulman, John and Zoph, Barret and Kim, Christina and Hilton, Jacob and Menick, Jacob and Weng, Jiayi and Uribe, Juan Felipe Ceron and Fedus, Liam and Metz, Luke and Pokorny, Michael and others},
  journal={OpenAI blog},
  volume={2},
  number={4},
  year={2022}
}

@article{lee2023rlaif,
  title={Rlaif vs. rlhf: Scaling reinforcement learning from human feedback with ai feedback},
  author={Lee, Harrison and Phatale, Samrat and Mansoor, Hassan and Mesnard, Thomas and Ferret, Johan and Lu, Kellie and Bishop, Colton and Hall, Ethan and Carbune, Victor and Rastogi, Abhinav and others},
  journal={arXiv preprint arXiv:2309.00267},
  year={2023}
}

@article{wang2022self,
  title={Self-consistency improves chain of thought reasoning in language models},
  author={Wang, Xuezhi and Wei, Jason and Schuurmans, Dale and Le, Quoc and Chi, Ed and Narang, Sharan and Chowdhery, Aakanksha and Zhou, Denny},
  journal={arXiv preprint arXiv:2203.11171},
  year={2022}
}

@article{xu2025softcot,
  title={Softcot: Soft chain-of-thought for efficient reasoning with llms},
  author={Xu, Yige and Guo, Xu and Zeng, Zhiwei and Miao, Chunyan},
  journal={arXiv preprint arXiv:2502.12134},
  year={2025}
}

@article{hao2024training,
  title={Training large language models to reason in a continuous latent space},
  author={Hao, Shibo and Sukhbaatar, Sainbayar and Su, DiJia and Li, Xian and Hu, Zhiting and Weston, Jason and Tian, Yuandong},
  journal={arXiv preprint arXiv:2412.06769},
  year={2024}
}

@inproceedings{yang2023diffusion,
  title={Diffusion model as representation learner},
  author={Yang, Xingyi and Wang, Xinchao},
  booktitle={Proceedings of the IEEE/CVF International Conference on Computer Vision},
  pages={18938--18949},
  year={2023}
}

@article{van2017neural,
  title={Neural discrete representation learning},
  author={Van Den Oord, Aaron and Vinyals, Oriol and others},
  journal={Advances in neural information processing systems},
  volume={30},
  year={2017}
}

@inproceedings{bowman2016generating,
  title={Generating sentences from a continuous space},
  author={Bowman, Samuel and Vilnis, Luke and Vinyals, Oriol and Dai, Andrew and Jozefowicz, Rafal and Bengio, Samy},
  booktitle={Proceedings of the 20th SIGNLL conference on computational natural language learning},
  pages={10--21},
  year={2016}
}

@article{he2025deepmath,
  title={Deepmath-103k: A large-scale, challenging, decontaminated, and verifiable mathematical dataset for advancing reasoning},
  author={He, Zhiwei and Liang, Tian and Xu, Jiahao and Liu, Qiuzhi and Chen, Xingyu and Wang, Yue and Song, Linfeng and Yu, Dian and Liang, Zhenwen and Wang, Wenxuan and others},
  journal={arXiv preprint arXiv:2504.11456},
  year={2025}
}

@article{he2019lagging,
  title={Lagging inference networks and posterior collapse in variational autoencoders},
  author={He, Junxian and Spokoyny, Daniel and Neubig, Graham and Berg-Kirkpatrick, Taylor},
  journal={arXiv preprint arXiv:1901.05534},
  year={2019}
}

@article{ho2020denoising,
  title={Denoising diffusion probabilistic models},
  author={Ho, Jonathan and Jain, Ajay and Abbeel, Pieter},
  journal={Advances in neural information processing systems},
  volume={33},
  pages={6840--6851},
  year={2020}
}

@article{kingma2013auto,
  title={Auto-encoding variational bayes},
  author={Kingma, Diederik P and Welling, Max},
  journal={arXiv preprint arXiv:1312.6114},
  year={2013}
}

@article{dathathri2019plug,
  title={Plug and play language models: A simple approach to controlled text generation},
  author={Dathathri, Sumanth and Madotto, Andrea and Lan, Janice and Hung, Jane and Frank, Eric and Molino, Piero and Yosinski, Jason and Liu, Rosanne},
  journal={arXiv preprint arXiv:1912.02164},
  year={2019}
}

@article{krause2020gedi,
  title={Gedi: Generative discriminator guided sequence generation},
  author={Krause, Ben and Gotmare, Akhilesh Deepak and McCann, Bryan and Keskar, Nitish Shirish and Joty, Shafiq and Socher, Richard and Rajani, Nazneen Fatema},
  journal={arXiv preprint arXiv:2009.06367},
  year={2020}
}

@article{turner2023steering,
  title={Steering language models with activation engineering},
  author={Turner, Alexander Matt and Thiergart, Lisa and Leech, Gavin and Udell, David and Vazquez, Juan J and Mini, Ulisse and MacDiarmid, Monte},
  journal={arXiv preprint arXiv:2308.10248},
  year={2023}
}

@article{fernando2023promptbreeder,
  title={Promptbreeder: Self-referential self-improvement via prompt evolution},
  author={Fernando, Chrisantha and Banarse, Dylan and Michalewski, Henryk and Osindero, Simon and Rockt{\"a}schel, Tim},
  journal={arXiv preprint arXiv:2309.16797},
  year={2023}
}

@article{shin2020autoprompt,
  title={Autoprompt: Eliciting knowledge from language models with automatically generated prompts},
  author={Shin, Taylor and Razeghi, Yasaman and Logan IV, Robert L and Wallace, Eric and Singh, Sameer},
  journal={arXiv preprint arXiv:2010.15980},
  year={2020}
}

@article{li2021prefix,
  title={Prefix-tuning: Optimizing continuous prompts for generation},
  author={Li, Xiang Lisa and Liang, Percy},
  journal={arXiv preprint arXiv:2101.00190},
  year={2021}
}

@article{lester2021power,
  title={The power of scale for parameter-efficient prompt tuning},
  author={Lester, Brian and Al-Rfou, Rami and Constant, Noah},
  journal={arXiv preprint arXiv:2104.08691},
  year={2021}
}

@article{yao2023tree,
  title={Tree of thoughts: Deliberate problem solving with large language models},
  author={Yao, Shunyu and Yu, Dian and Zhao, Jeffrey and Shafran, Izhak and Griffiths, Tom and Cao, Yuan and Narasimhan, Karthik},
  journal={Advances in neural information processing systems},
  volume={36},
  pages={11809--11822},
  year={2023}
}

@inproceedings{yao2022react,
  title={React: Synergizing reasoning and acting in language models},
  author={Yao, Shunyu and Zhao, Jeffrey and Yu, Dian and Du, Nan and Shafran, Izhak and Narasimhan, Karthik R and Cao, Yuan},
  booktitle={The eleventh international conference on learning representations},
  year={2022}
}

@article{zhou2022least,
  title={Least-to-most prompting enables complex reasoning in large language models},
  author={Zhou, Denny and Sch{\"a}rli, Nathanael and Hou, Le and Wei, Jason and Scales, Nathan and Wang, Xuezhi and Schuurmans, Dale and Cui, Claire and Bousquet, Olivier and Le, Quoc and others},
  journal={arXiv preprint arXiv:2205.10625},
  year={2022}
}

@article{wei2022chain,
  title={Chain-of-thought prompting elicits reasoning in large language models},
  author={Wei, Jason and Wang, Xuezhi and Schuurmans, Dale and Bosma, Maarten and Xia, Fei and Chi, Ed and Le, Quoc V and Zhou, Denny and others},
  journal={Advances in neural information processing systems},
  volume={35},
  pages={24824--24837},
  year={2022}
}

@article{li2025attention,
  title={Attention Illuminates LLM Reasoning: The Preplan-and-Anchor Rhythm Enables Fine-Grained Policy Optimization},
  author={Li, Yang and Dong, Zhichen and Sun, Yuhan and Wang, Weixun and Xiong, Shaopan and Luo, Yijia and Liu, Jiashun and Lu, Han and Wang, Jiamang and Su, Wenbo and others},
  journal={arXiv preprint arXiv:2510.13554},
  year={2025}
}

@misc{bai2022constitutionalaiharmlessnessai,
      title={Constitutional AI: Harmlessness from AI Feedback}, 
      author={Yuntao Bai and Saurav Kadavath and Sandipan Kundu and Amanda Askell and Jackson Kernion and Andy Jones and Anna Chen and Anna Goldie and Azalia Mirhoseini and Cameron McKinnon and Carol Chen and Catherine Olsson and Christopher Olah and Danny Hernandez and Dawn Drain and Deep Ganguli and Dustin Li and Eli Tran-Johnson and Ethan Perez and Jamie Kerr and Jared Mueller and Jeffrey Ladish and Joshua Landau and Kamal Ndousse and Kamile Lukosuite and Liane Lovitt and Michael Sellitto and Nelson Elhage and Nicholas Schiefer and Noemi Mercado and Nova DasSarma and Robert Lasenby and Robin Larson and Sam Ringer and Scott Johnston and Shauna Kravec and Sheer El Showk and Stanislav Fort and Tamera Lanham and Timothy Telleen-Lawton and Tom Conerly and Tom Henighan and Tristan Hume and Samuel R. Bowman and Zac Hatfield-Dodds and Ben Mann and Dario Amodei and Nicholas Joseph and Sam McCandlish and Tom Brown and Jared Kaplan},
      year={2022},
      eprint={2212.08073},
      archivePrefix={arXiv},
      primaryClass={cs.CL},
      url={https://arxiv.org/abs/2212.08073}, 
}

@misc{guo2024directlanguagemodelalignment,
      title={Direct Language Model Alignment from Online AI Feedback}, 
      author={Shangmin Guo and Biao Zhang and Tianlin Liu and Tianqi Liu and Misha Khalman and Felipe Llinares and Alexandre Rame and Thomas Mesnard and Yao Zhao and Bilal Piot and Johan Ferret and Mathieu Blondel},
      year={2024},
      eprint={2402.04792},
      archivePrefix={arXiv},
      primaryClass={cs.AI},
      url={https://arxiv.org/abs/2402.04792}, 
}

@article{sun2023principle,
  title={Principle-driven self-alignment of language models from scratch with minimal human supervision},
  author={Sun, Zhiqing and Shen, Yikang and Zhou, Qinhong and Zhang, Hongxin and Chen, Zhenfang and Cox, David and Yang, Yiming and Gan, Chuang},
  journal={Advances in Neural Information Processing Systems},
  volume={36},
  pages={2511--2565},
  year={2023}
}

@article{bai2022training,
  title={Training a helpful and harmless assistant with reinforcement learning from human feedback},
  author={Bai, Yuntao and Jones, Andy and Ndousse, Kamal and Askell, Amanda and Chen, Anna and DasSarma, Nova and Drain, Dawn and Fort, Stanislav and Ganguli, Deep and Henighan, Tom and others},
  journal={arXiv preprint arXiv:2204.05862},
  year={2022}
}

@misc{lambert2025tulu3pushingfrontiers,
      title={Tulu 3: Pushing Frontiers in Open Language Model Post-Training}, 
      author={Nathan Lambert and Jacob Morrison and Valentina Pyatkin and Shengyi Huang and Hamish Ivison and Faeze Brahman and Lester James V. Miranda and Alisa Liu and Nouha Dziri and Shane Lyu and Yuling Gu and Saumya Malik and Victoria Graf and Jena D. Hwang and Jiangjiang Yang and Ronan Le Bras and Oyvind Tafjord and Chris Wilhelm and Luca Soldaini and Noah A. Smith and Yizhong Wang and Pradeep Dasigi and Hannaneh Hajishirzi},
      year={2025},
      eprint={2411.15124},
      archivePrefix={arXiv},
      primaryClass={cs.CL},
      url={https://arxiv.org/abs/2411.15124}, 
}

@misc{xin2024deepseekproverv15harnessingproofassistant,
      title={DeepSeek-Prover-V1.5: Harnessing Proof Assistant Feedback for Reinforcement Learning and Monte-Carlo Tree Search}, 
      author={Huajian Xin and Z. Z. Ren and Junxiao Song and Zhihong Shao and Wanjia Zhao and Haocheng Wang and Bo Liu and Liyue Zhang and Xuan Lu and Qiushi Du and Wenjun Gao and Qihao Zhu and Dejian Yang and Zhibin Gou and Z. F. Wu and Fuli Luo and Chong Ruan},
      year={2024},
      eprint={2408.08152},
      archivePrefix={arXiv},
      primaryClass={cs.CL},
      url={https://arxiv.org/abs/2408.08152}, 
}

@misc{wang2024mathshepherdverifyreinforcellms,
      title={Math-Shepherd: Verify and Reinforce LLMs Step-by-step without Human Annotations}, 
      author={Peiyi Wang and Lei Li and Zhihong Shao and R. X. Xu and Damai Dai and Yifei Li and Deli Chen and Y. Wu and Zhifang Sui},
      year={2024},
      eprint={2312.08935},
      archivePrefix={arXiv},
      primaryClass={cs.AI},
      url={https://arxiv.org/abs/2312.08935}, 
}

@misc{zheng2025groupsequencepolicyoptimization,
      title={Group Sequence Policy Optimization}, 
      author={Chujie Zheng and Shixuan Liu and Mingze Li and Xiong-Hui Chen and Bowen Yu and Chang Gao and Kai Dang and Yuqiong Liu and Rui Men and An Yang and Jingren Zhou and Junyang Lin},
      year={2025},
      eprint={2507.18071},
      archivePrefix={arXiv},
      primaryClass={cs.LG},
      url={https://arxiv.org/abs/2507.18071}, 
}

@article{wang2025reinforcement,
  title={Reinforcement Learning Optimization for Large-Scale Learning: An Efficient and User-Friendly Scaling Library},
  author={Wang, Weixun and Xiong, Shaopan and Chen, Gengru and Gao, Wei and Guo, Sheng and He, Yancheng and Huang, Ju and Liu, Jiaheng and Li, Zhendong and Li, Xiaoyang and others},
  journal={arXiv preprint arXiv:2506.06122},
  year={2025}
}

@article{cobbe2021gsm8k,
  title={Training Verifiers to Solve Math Word Problems},
  author={Cobbe, Karl and Kosaraju, Vineet and Bavarian, Mohammad and Chen, Mark and Jun, Heewoo and Kaiser, Lukasz and Plappert, Matthias and Tworek, Jerry and Hilton, Jacob and Nakano, Reiichiro and Hesse, Christopher and Schulman, John},
  journal={arXiv preprint arXiv:2110.14168},
  year={2021}
}
}


\end{document}